\documentclass[12pt, a4paper]{article}

\usepackage{PRIMEarxiv}
\usepackage{enumitem}
\setitemize{topsep=0pt, leftmargin=*}
\usepackage[utf8]{inputenc}
\usepackage{array}
\usepackage{adjustbox}
\usepackage{amsmath,amsfonts,amssymb}
\usepackage{amsthm}
\usepackage{algorithm,algorithmic}
\usepackage{booktabs}
\usepackage{graphicx}
\usepackage{float}
\usepackage{fancyhdr}
\usepackage{caption}
\usepackage{subcaption}
\usepackage{hyperref}
\usepackage{url}
\usepackage{mathtools}
\usepackage{bm}
\usepackage{multirow} 
\usepackage[nameinlink]{cleveref}
\usepackage{tikz}
\usetikzlibrary{shapes.geometric,arrows,positioning,fit}

\usepackage{pdflscape}  
\usepackage{lscape}     

\usepackage[numbers,sort&compress]{natbib}

\newtheorem{Theorem}{Theorem}
\newtheorem{Lemma}{Lemma}
\newtheorem{definition}{Definition}
\newtheorem{Proposition}{Proposition}

\title{Personalized Federated Heat-Kernel Enhanced Multi-View Clustering via Advanced Tensor Decomposition Techniques}
\author{
    Kristina P. Sinaga\thanks{Corresponding author: Independent Researcher. ORCID: 0009-0000-6184-829X. Email: kristinasinaga41@gmail.com. Research interests include federated learning, multi-view clustering, privacy-preserving machine learning, heat kernel methods, quantum field theory applications in machine learning, and distributed intelligent systems.}\\
    \small Independent Researcher\\
    \small \texttt{kristinasinaga41@gmail.com}
}
\date{January 1st, 2026}

\begin{document}
\maketitle
\vspace{-1cm}
\begin{figure}[H]
\centering
\begin{tikzpicture}[
    node distance=0.8cm and 1.2cm,
    box/.style={rectangle, draw, rounded corners, minimum width=2.2cm, minimum height=0.7cm, align=center, font=\scriptsize},
    clientbox/.style={box, fill=blue!15, draw=blue!60},
    serverbox/.style={box, fill=green!15, draw=green!60},
    processbox/.style={box, fill=orange!15, draw=orange!60},
    tensorbox/.style={box, fill=purple!15, draw=purple!60},
    personbox/.style={box, fill=red!15, draw=red!60},
    arrow/.style={->, >=stealth, thick},
    dashedarrow/.style={->, >=stealth, thick, dashed},
    label/.style={font=\tiny, align=center}
]

\node[font=\small\bfseries] at (0,5.5) {Unified FedHK Framework for Multi-View Clustering};

\node[font=\scriptsize\bfseries, blue!70] at (-5.5,4.5) {Clients $\ell = 1, \ldots, M$};

\node[clientbox] (mvdata) at (-5.5,3.5) {Multi-View Data\\$X_{[\ell]}^h, h=1,\ldots,s$};

\node[tensorbox] (tensor) at (-5.5,2.3) {Tensorization\\$\mathcal{X}_{[\ell]} \in \mathbb{R}^{n \times D \times s}$};

\node[processbox] (hkc) at (-5.5,1.1) {Heat-Kernel\\Coefficients $\delta_{[\ell]ij}^h$};

\node[font=\tiny, fill=gray!20, draw, circle, inner sep=1pt] (branch) at (-5.5,-0.1) {};

\node[clientbox] (efkmvc) at (-7.5,-1.2) {E-FKMVC\\FKED Distance};

\node[tensorbox] (decomp) at (-3.5,-1.2) {Tensor Decomposition};

\node[tensorbox] (parafac) at (-4.8,-2.4) {PARAFAC2\\$\mathcal{G}, \mathbf{P}, \mathbf{Q}, \mathbf{R}$};
\node[tensorbox] (tucker) at (-2.2,-2.4) {Tucker\\$\mathcal{G}, \mathbf{U}, \mathbf{V}, \mathbf{W}$};

\node[processbox] (localclust) at (-6.1,-3.6) {Local Clustering\\Update $U^*, A, V$};

\node[font=\scriptsize\bfseries, green!70] at (3,4.5) {Central Server};

\node[serverbox] (agg) at (3,2.8) {Privacy-Preserving\\Aggregation};

\node[serverbox] (global) at (3,1.4) {Global Model\\$\Theta_{\text{global}}$};

\node[personbox] (person) at (3,0) {Personalization\\$\lambda_\ell, \rho_\ell$ Adaptation};

\node[serverbox] (broadcast) at (3,-1.4) {Broadcast to\\Clients};

\node[processbox] (stats) at (-0.8,-3.6) {Shared Statistics\\$S_{[\ell]}^{centers}, S_{[\ell]}^{views}$};

\draw[arrow] (mvdata) -- (tensor);
\draw[arrow] (mvdata) to[out=-90, in=180] (hkc);
\draw[arrow] (tensor) -- (hkc);
\draw[arrow] (hkc) -- (branch);
\draw[arrow] (branch) -- (-7.5,-0.1) -- (efkmvc);
\draw[arrow] (branch) -- (-3.5,-0.1) -- (decomp);
\draw[arrow] (decomp) -- (parafac);
\draw[arrow] (decomp) -- (tucker);
\draw[arrow] (efkmvc) -- (-7.5,-3.6) -- (localclust);
\draw[arrow] (parafac) -- (-4.8,-3.6) -- (localclust);
\draw[arrow] (tucker) -- (-2.2,-3.6) -- (localclust);
\draw[arrow] (localclust) -- (stats);

\draw[arrow, blue!60] (stats) -- (1,0) -- (1,2.8) -- (agg);

\draw[arrow] (agg) -- (global);
\draw[arrow] (global) -- (person);
\draw[arrow] (person) -- (broadcast);

\draw[dashedarrow, green!60] (broadcast) -- (1,-1.4) -- (1,-4.5) -- (-5.5,-4.5) -- (-5.5,-4);
\node[label] at (-2,-4.7) {Global updates + Personalization};

\draw[dashedarrow, red!60] (-5.5,-4) -- (-5.5,-3.8);

\node[font=\tiny, fill=blue!10, draw=blue!40, rounded corners] at (-7.5,-0.4) {E-FKMVC};
\node[font=\tiny, fill=purple!10, draw=purple!40, rounded corners] at (-5.4,-1.8) {FedHK-PARAFAC2};
\node[font=\tiny, fill=purple!10, draw=purple!40, rounded corners] at (-1.8,-1.8) {FedHK-Tucker};
\node[font=\tiny, fill=red!10, draw=red!40, rounded corners] at (3,-0.7) {FedHK-MVC-Person};

\node[font=\tiny\bfseries] at (5.5,3.5) {Legend:};
\node[clientbox, minimum width=1.2cm, minimum height=0.4cm] at (5.5,3) {};
\node[font=\tiny, right] at (6.2,3) {Client};
\node[serverbox, minimum width=1.2cm, minimum height=0.4cm] at (5.5,2.5) {};
\node[font=\tiny, right] at (6.2,2.5) {Server};
\node[tensorbox, minimum width=1.2cm, minimum height=0.4cm] at (5.5,2) {};
\node[font=\tiny, right] at (6.2,2) {Tensor};
\node[personbox, minimum width=1.2cm, minimum height=0.4cm] at (5.5,1.5) {};
\node[font=\tiny, right] at (6.2,1.5) {Personal};

\end{tikzpicture}
\caption{Graphical Abstract: Unified Framework for Federated Heat-Kernel Enhanced Multi-View Clustering. The framework integrates four algorithms: E-FKMVC (direct matrix-based), FedHK-PARAFAC2 and FedHK-Tucker (tensor decomposition-based), and FedHK-MVC-Person (personalized aggregation). Clients perform local clustering with heat-kernel coefficients and communicate privacy-preserving statistics to the server for global aggregation and personalized model updates.}
\label{fig:graphical_abstract}
\end{figure}

\begin{abstract}
This paper introduces mathematical frameworks that address the challenges of multi-view clustering in federated learning environments. The objective is to integrate optimization techniques based on new objective functions employing heat-kernel coefficients to replace conventional distance metrics with quantum-inspired measures. The proposed frameworks utilize advanced tensor decomposition methods—specifically, PARAFAC2 and Tucker decomposition—to efficiently represent high-dimensional, multi-view data while preserving inter-view relationships. The research has yielded the development of four novel algorithms: an efficient federated kernel multi-view clustering (E-FKMVC) model; FedHK-PARAFAC2; FedHK-Tucker; and FedHK-MVC-Person with PARAFAC2 Decomposition (Personalized FedHK-PARAFAC2). The primary objective of these algorithms is to enhance the efficacy of clustering processes while ensuring the confidentiality and efficient communication in federated learning environments. Theoretical analyses of convergence guarantees, privacy bounds, and complexity are provided to validate the effectiveness of the proposed methods. In essence, this paper makes a significant academic contribution to the field of federated multi-view clustering through its innovative integration of mathematical modeling and algorithm design. This approach addresses the critical challenges of data heterogeneity and privacy concerns, paving the way for enhanced data management and analytics in various contexts.

\end{abstract}

\section{Introduction}
\label{sec:Introduction}

Large-scale data centers become increasingly important as the Internet of Things (IoT) continues to grow, with more devices being massively connected to the internet than ever before. This phenomenal transition of data to distributed environments affects the need for investigating optimal approaches to process heterogeneous information from multiple devices while ensuring security and efficient data exchange. Conventional unsupervised techniques cannot effectively handle these higher-order or higher-dimensional data structures inherent in modern federated systems where distributed clients possess different data modalities, feature dimensions, and local patterns. 

\begin{figure}[ht!]
\centering
\includegraphics[width=0.5\textwidth]{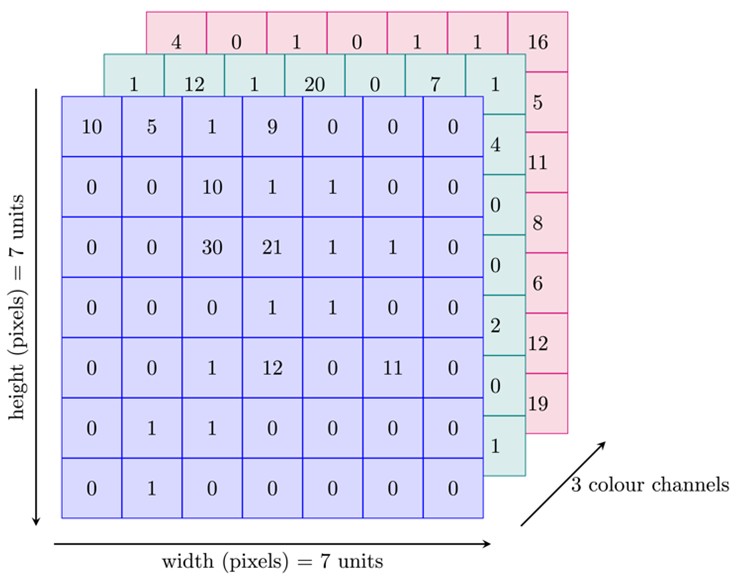}
\caption{The Illustration of a Color Image as a Third-Order Tensor Representation with RGB Channels. The channel dimension captures the color information, while the height and width dimensions represent the spatial structure of the image.}
\label{fig:RGBthirdorder3DImage}
\end{figure}

Tensor-based machine learning approaches have emerged as powerful tools for analyzing multi-dimensional data structures, enabling the capture of complex relationships and interactions across multiple modes. Tensors, as generalizations of matrices to higher dimensions, provide a natural framework for representing multi-view data, where each view corresponds to a different modality or feature set. For instance, in large language models (LLMs), text data can be represented as a three-dimensional tensor, with dimensions corresponding to words, contexts, and semantic features. This tensor representation allows for the preservation of contextual information and facilitates the extraction of meaningful patterns through tensor decomposition techniques. In computer vision, images and videos can be represented as higher-order tensors, capturing spatial and temporal relationships across multiple channels. For example, a color image can be represented as a third-order tensor with dimensions corresponding to height, width, and color channels (RGB). As illustrated in Figure \ref{fig:RGBthirdorder3DImage}, a color image is represented as a third-order tensor, where each slice along the third dimension corresponds to one of the RGB color channels. This tensor representation enables the preservation of spatial and color information, facilitating more effective analysis and processing of image data.

Multi-view data can be effectively represented as higher-order tensors, capturing the complex relationships and interactions across different views. Each view corresponds to a different modality or feature set, and the tensor representation allows for the preservation of inter-view relationships. As demonstrated in Figure \ref{fig:tensor_representation_mv_data}, a color image dataset with multiple views, such as RGB, depth, infrared (IR), and thermal channels, can be represented as a fourth-order tensor. Each slice along the fourth dimension corresponds to a different view of the same image, capturing complementary information across modalities. This tensor representation enables the application of advanced tensor decomposition techniques to extract shared and view-specific features, facilitating more effective multi-view data analysis.

In federated learning environments, clients often possess multi-view (MV) data that can be naturally represented as tensors. For example, in healthcare applications, different hospitals may collect various types of patient data, such as medical images, electronic health records, and genomic data, each representing a different view of the patient's health status. By leveraging tensor-based representations, federated learning algorithms can effectively capture the multi-modal nature of the data while preserving privacy and minimizing communication overhead. Singular value decomposition (SVD) and principal component analysis (PCA) \cite{welke2006fast, sinaga2021entropy} are classical matrix factorization techniques that have been widely used for dimensionality reduction and feature extraction. However, these methods are limited to two-dimensional data structures and cannot fully exploit the multi-way relationships present in tensor data. Tensor decomposition methods, such as CANDECOMP/PARAFAC (CP) \cite{harshman1970foundations} and Tucker decomposition \cite{tucker1966some}, extend these concepts to higher-order tensors, enabling more comprehensive analysis of multi-view data.

From the perspective of MV clustering, each component of feature-view stored as a \textit{two-order} tensor (matrix) can be naturally represented as a \textit{three-order} tensor by stacking the matrices along a new dimension corresponding to the views. This tensor representation allows for the preservation of inter-view relationships and facilitates the application of tensor decomposition techniques to extract shared and view-specific features. By leveraging tensor algebra, MV clustering algorithms can effectively capture the complex interactions between different views, leading to improved clustering performance and robustness. 

\begin{figure}[!htbp]
\centering
\begin{tikzpicture}[scale=1.2]
    \definecolor{view1}{RGB}{255,100,100}  
    \definecolor{view2}{RGB}{100,255,100}  
    \definecolor{view3}{RGB}{100,100,255}  
    \definecolor{view4}{RGB}{255,200,100}  

    \fill[view4, opacity=0.3] (0.9,0.9,0) -- (3.9,0.9,0) -- (3.9,3.9,0) -- (0.9,3.9,0) -- cycle;
    \draw[thick, view4!70!black] (0.9,0.9,0) -- (3.9,0.9,0) -- (3.9,3.9,0) -- (0.9,3.9,0) -- cycle;

    \fill[view3, opacity=0.4] (0.6,0.6,0.8) -- (3.6,0.6,0.8) -- (3.6,3.6,0.8) -- (0.6,3.6,0.8) -- cycle;
    \draw[thick, view3!70!black] (0.6,0.6,0.8) -- (3.6,0.6,0.8) -- (3.6,3.6,0.8) -- (0.6,3.6,0.8) -- cycle;

    \fill[view2, opacity=0.5] (0.3,0.3,1.6) -- (3.3,0.3,1.6) -- (3.3,3.3,1.6) -- (0.3,3.3,1.6) -- cycle;
    \draw[thick, view2!70!black] (0.3,0.3,1.6) -- (3.3,0.3,1.6) -- (3.3,3.3,1.6) -- (0.3,3.3,1.6) -- cycle;

    \fill[view1, opacity=0.6] (0,0,2.4) -- (3,0,2.4) -- (3,3,2.4) -- (0,3,2.4) -- cycle;
    \draw[thick, view1!70!black] (0,0,2.4) -- (3,0,2.4) -- (3,3,2.4) -- (0,3,2.4) -- cycle;

    \draw[thick, gray, dashed] (0,0,2.4) -- (0.9,0.9,0);
    \draw[thick, gray, dashed] (3,0,2.4) -- (3.9,0.9,0);
    \draw[thick, gray, dashed] (3,3,2.4) -- (3.9,3.9,0);
    \draw[thick, gray, dashed] (0,3,2.4) -- (0.9,3.9,0);

    \node[font=\small] at (1.5,-0.5,2.4) {Width};
    \node[font=\small, rotate=90] at (-0.6,1.5,2.4) {Height};
    \node[font=\small] at (5,4.5,1.2) {Views};

    \node[view1, fill=view1!30, rounded corners, font=\small] at (3.2,2.2,2.4) {View 1 (RGB)};
    \node[view2!70!black, fill=view2!30, rounded corners, font=\small] at (3.8,2.6,1.6) {View 2 (Depth)};
    \node[view3!70!black, fill=view3!30, rounded corners, font=\small] at (4.1,3,0.8) {View 3 (IR)};
    \node[view4!70!black, fill=view4!30, rounded corners, font=\small] at (5,3.4,0) {View 4 (Thermal)};

    \draw[->, thick, purple] (4.5,0.5,2.4) -- (5.3,1.3,0);
    \node[purple, font=\footnotesize] at (5.5,0.5,1.2) {4th mode};

\end{tikzpicture}
\caption{Tensor Representation of Multi-View Data with Different Modalities. Each slice along the fourth dimension corresponds to a different view of the same image, capturing complementary information across modalities such as RGB, depth, infrared (IR), and thermal channels.}
\label{fig:tensor_representation_mv_data}
\end{figure}

The analogy of diversity and heterogeneity on multiple representations of the same information is precise \cite{giunchiglia2022representation} . In this sense, multi-view data can be viewed as different perspectives or modalities of the same underlying phenomenon, each providing unique insights and complementary information. For example, in social network analysis, user profiles may include text data from posts, images shared, and interaction patterns with other users. Each of these views captures different aspects of user behavior and preferences, contributing to a more holistic understanding of the social dynamics. 

This paper presents new frameworks that integrate heat-kernel coefficients and multi-view fuzzy c-means clustering via advanced tensor decomposition techniques. The framework's scope extends beyond the realm of addressing challenges related to data heterogeneity, privacy preservation, and communication efficiency in federated learning environments. Furthermore, it overcomes the limitations of existing MV clustering methods in both federated and non-federated settings, thereby addressing the issue of data decomposition in MV clustering that does not employ deep learning. The contribution of this paper is threefold. 
\begin{itemize}
    \item Four novel algorithms are presented, based on a basic, efficient federated kernel multi-view clustering (E-FKMVC) model. This model integrates the heat-kernel method to replace the conventional Euclidean distance metric with heat-kernel coefficients adapted from quantum field theory. The FedHK-PARAFAC2, FedHK-Tucker, and Personalized FedHK-PARAFAC2/Tucker algorithms are based on the E-FKMVC objective function. 
    \item The integration of tensor decomposition techniques, specifically PARAFAC2 and Tucker decomposition, into the federated multi-view clustering framework to efficiently represent high-dimensional multi-view structures while preserving inter-view relationships.
    \item Theoretical analysis of the proposed algorithms, including convergence guarantees, privacy bounds, and complexity analysis, to ensure their effectiveness and efficiency in federated learning environments.
\end{itemize}

The subsequent sections of this paper are organized as follows. In section \ref{sec:Related_Work}, the extant literature on federated learning and clustering, multi-view clustering, heat kernel methods and quantum-inspired learning, and tensor decomposition techniques is reviewed. In Section \ref{sec:Mathematical_Preliminaries}, the reader will find concise theoretical and mathematical backgrounds on federated learning and clustering, multi-view clustering, heat kernel methods, and tensor decomposition. The proposed efficient federated kernel multi-view clustering (E-FKMVC) framework is delineated in Section \ref{sec:proposed_methodology}. As delineated in Section \ref{sec:proposed_fed_tensorized_clustering}, the proposed FedHK-PARAFAC2, FedHK-Tucker, and its personalized algorithms are described. In accordance with the findings of the research, Section \ref{sec:conclusion} of the paper presents a conclusion. Additionally, it delineates the potential avenues for future investigation and the practical applications of the research.

\section{Related Work}
\label{sec:Related_Work}

In order to provide a comprehensive overview of the contributions of this study, a review of the extant literature was conducted across four key domains: federated learning and clustering, multi-view clustering, heat kernel methods and quantum-inspired learning, and tensor decomposition techniques. This review underscores foundational works and recent advancements in the field, while concurrently identifying gaps that the proposed framework aims to address. 

\subsection{Federated Learning and Clustering}

Pattern recognition and clustering in federated learning (FL) environments have garnered significant attention due to the increasing need for privacy-preserving distributed data analysis. Generally, federated learning can be categorized into two types: horizontal federated learning (HFL) and vertical federated learning (VFL) (See Riviera et al. (2025) \cite{11000331} for a comprehensive survey on FL). HFL is applicable when datasets across clients share the same feature space but differ in samples, while VFL is suitable when datasets share the same sample space but differ in features. Early work by McMahan et al. (2017) \cite{mcmahan2017communication} introduced the Federated Averaging (FedAvg) algorithm, which laid the groundwork for subsequent FL research. Kone{\v{c}}n{\`y} et al. (2016) \cite{konevcny2016federated} introduced federated optimization as a method to train a centralized model using decentralized data stored across numerous devices, emphasizing communication efficiency due to limited connectivity and privacy concerns. They proposed modifications to existing algorithms to adapt them to federated optimization settings. Geyer et al. (2017) \cite{geyer2017differentially} demonstrated how differential privacy can be integrated into federated clustering while maintaining acceptable utility. Li et al. (2020) \cite{li2020acceleration} adopted the foundational ideas of federated optimization and extended them by integrating advanced techniques like gradient compression, acceleration, and variance reduction to address communication bottlenecks and improve the scalability of federated learning systems. However, these privacy mechanisms often come at the cost of accuracy, particularly in high-dimensional settings. In the context of clustering, several approaches have been proposed to adapt traditional clustering algorithms to federated settings. For instance, Dennis et al. (2021) \cite{dennis2021heterogeneity} explored federated k-means clustering, demonstrating the feasibility of distributed clustering with local updates and global aggregation. Ghosh et al. (2002) \cite{ghosh2002consensus} proposed consensus-based approaches for federated hierarchical clustering, addressing challenges of data heterogeneity and communication efficiency. Makhija et al. (2022) \cite{makhija2022federated} introduced federated fuzzy c-means clustering, leveraging soft assignments to enhance robustness against noise and outliers. However, these methods often assume homogeneous data distributions and do not fully exploit multi-view data structures prevalent in modern applications. Recent advancements have focused on personalized federated clustering to accommodate client heterogeneity. Caldarola et al. (2021) \cite{caldarola2021cluster} proposed adaptive weight schemes for heterogeneous clustering, while Li et al. (2022) \cite{li2022hpfl} developed meta-learning approaches for personalized cluster initialization. Privacy-preserving aspects have been addressed through differential privacy mechanisms \cite{kairouz2021advances} and secure aggregation protocols \cite{bonawitz2017practical}, ensuring that sensitive information remains protected during the clustering process. Communication efficiency has been tackled through gradient compression \cite{konecny2016federated}, sparsification techniques \cite{aji2017sparse}, and model quantization \cite{reisizadeh2020fedpaq}, reducing the overhead associated with transmitting model updates across clients. However, these methods often face challenges related to data heterogeneity, communication efficiency, and convergence guarantees in distributed environments.

\subsection{Multi-View Clustering}

Multi-view clustering (MVC) has emerged as a powerful paradigm for leveraging complementary information from multiple data representations to enhance clustering performance. Early work by Bickel and Scheffer (2004) \cite{bickel2004multi} established the theoretical foundations for multi-view learning, demonstrating that integrating multiple views can lead to improved clustering accuracy. Classical MVC approaches can be categorized into co-training based methods, subspace learning approaches, and consensus clustering techniques. Co-training methods, such as those proposed by Kumar et al. (2011) \cite{kumar2011co}, iteratively refine clustering results by leveraging inter-view consistency. Subspace learning approaches, including Canonical Correlation Analysis (CCA) based methods \cite{hotelling1992relations} and shared subspace clustering \cite{wang2019multiple}, project multiple views into a common representation space for unified clustering. Graph-based MVC has gained significant attention due to its ability to capture complex inter-view relationships. Wang et al. (2019) \cite{wang2019gmc} proposed multi-view graph clustering with adaptive neighbor learning, while Nie et al. (2016) \cite{nie2016parameter} developed parameter-free auto-weighted multiple graph learning. Deep learning approaches have revolutionized MVC through end-to-end representation learning. Zhang et al. (2019)  \cite{zhang2019ae2} introduced auto-encoder based multi-view clustering, while Zhu et al. (2019) \cite{zhu2019multi} developed deep multi-view clustering networks with view correlation discovery. Fuzzy clustering extensions to multi-view scenarios have been explored by Jiang et al. (2014) \cite{jiang2014collaborative} and Wang et al. (2017) \cite{wang2017multi}, introducing weighted aggregation schemes and view-specific fuzzification parameters. Sinaga et al. (2025) \cite{sinaga2025parameter} proposed a parameter-free multi-view fuzzy clustering algorithm that automatically learns view weights and fuzzification parameters, enhancing clustering robustness and interpretability. However, these methods typically require centralized data access and are not designed for federated environments. 

The federated multi-view clustering is first introduced by Chen et al. (2022) \cite{chen2022federated} for federated multi-view representation learning and Liu et al. (2023) \cite{liu2023federated} for privacy-preserving multi-view clustering. However, these methods do not leverage tensor decomposition for efficient representation and communication. Yang et al. (2024) \cite{yang2024federated} proposed a federated multi-view clustering framework that combines local clustering with global consensus, but it lacks the integration of heat-kernel methods and advanced tensor decomposition techniques. Earlier, Sinaga (2025) \cite{sinaga2025fedhk} proposed a heat-kernel enhanced federated multi-view clustering framework, but it did not incorporate tensor decomposition methods for efficient multi-view representation. Thus, there is a need for advanced federated multi-view clustering approaches that leverage tensor decomposition and heat-kernel methods to address data heterogeneity, privacy preservation, and communication efficiency challenges in distributed settings. 

\subsection{Heat Kernel Methods and Quantum-Inspired Learning}

Heat kernel methods have emerged as powerful tools for capturing local geometric structure in high-dimensional data spaces. Originally developed in differential geometry and quantum field theory, heat kernels provide a natural way to encode multi-scale geometric information through diffusion processes on manifolds. The heat kernel, defined as the fundamental solution to the heat equation, has found extensive applications in machine learning through kernel methods and manifold learning. Ham et al. (2004) \cite{ham2004kernel} introduced diffusion maps that leverage heat kernel properties for dimensionality reduction, while Coifman and Lafon (2006)  \cite{coifman2006diffusion} established theoretical foundations for diffusion-based data analysis. In clustering applications, heat kernel methods have been used to capture local neighborhood structure and improve cluster boundary detection. Zelnik-Manor and Perona (2004) \cite{zelnik2004self} developed self-tuning spectral clustering using local heat kernel scales, while von Luxburg (2007) \cite{von2007tutorial} provided comprehensive analysis of heat kernel properties in graph-based clustering. 

Quantum-inspired machine learning has gained attention for its ability to leverage quantum mechanical principles in classical algorithms. Quantum kernels, introduced by Havlíček et al. (2019) \cite{havlivcek2019supervised}, enable the computation of quantum feature maps for enhanced pattern recognition. The connection between quantum field theory and machine learning has been explored through path integral formulations \cite{carleo2017solving} and quantum neural networks \cite{biamonte2017quantum}. The integration of heat kernel coefficients with clustering algorithms represents a natural extension of kernel methods to soft clustering scenarios. By incorporating heat diffusion properties, clustering algorithms can better capture local data manifold structure and improve robustness to noise and outliers. 

\subsection{Tensor Decomposition Methods}

Tensor decomposition has emerged as a powerful paradigm for analyzing high-dimensional multi-modal data, providing compact representations that capture multilinear relationships inherent in complex datasets. The field has evolved from classical matrix factorization techniques to sophisticated tensor algebra frameworks capable of handling arbitrary $N-way$ data structures.  Canonical polyadic decomposition (CPD) or parallel factors (PARAFAC) \cite{harshman1994parafac, carroll1970analysis}, Tucker decomposition (TD), and non-negative tensor factorization (NTF) are well-known approaches to reorder a tensor into a matrix or vector. PARAFAC or CPD holds multilinearity assumptions such as handle a sum of rank-one tensors by targeting the same lengths. In this sense, the original PARAFAC decomposition can reveal uncorrelated (orthogonal) and correlated (oblique) factors in system-variation data matrices, but failed to describe correlated factors when applied to the cross-product matrices computed from one data \cite{harshman1972parafac2, kiers1999parafac2}. PARAFAC2, on the other hand has the advantageous to handle irregular tensors by relaxing the canonical polyadic (CP) model \cite{harshman1972parafac2}. PARAFAC2 enables the three-mode model, often implemented into subspace clustering algorithms taking advantageous of correlations between pair of factors. The PARAFAC2 normalize the generalization of PARAFAC to deal both with uncorrelated and correlated factors.  Canonical Polyadic Decomposition (CPD), also known as CANDECOMP/PARAFAC, represents the foundational tensor decomposition method. Introduced independently by Carroll and Chang (1970) \cite{carroll1970analysis} and Harshman \cite{harshman1970foundations}, CPD decomposes a tensor into a sum of rank-one tensors, providing a unique factorization under mild conditions. Tucker decomposition, proposed by Tucker \cite{tucker1966some}, provides a more flexible alternative through its core tensor and factor matrices structure. Unlike CPD, Tucker decomposition allows different ranks along each mode, making it suitable for data with varying complexity across dimensions. The Higher-Order SVD (HOSVD) algorithm by De Lathauwer et al. \cite{de2000multilinear} established efficient computational procedures for Tucker decomposition, enabling its application to large-scale problems. 

Recent advances in tensor decomposition have focused on addressing practical challenges such as missing data, noise robustness, and computational scalability. Tensor completion methods, including work by Liu et al. \cite{liu2012tensor} and Yuan and Zhang \cite{yuan2016tensor}, enable handling of incomplete tensors through low-rank assumptions. Robust tensor decomposition techniques \cite{huang2014robust} provide resilience against outliers and corrupted entries. Non-negative tensor factorization (NTF) has gained attention for applications requiring interpretable, parts-based representations. Lee and Seung's pioneering work \cite{lee2000algorithms} on non-negative matrix factorization was extended to tensors by Shashua and Hazan \cite{shashua2005non}, enabling decompositions with non-negativity constraints that often yield more interpretable factors. The integration of tensor decomposition with machine learning has led to tensorized neural networks \cite{novikov2015tensorizing}, tensor-based kernel methods \cite{signoretto2011kernel}, and multi-task learning frameworks \cite{wimalawarne2014multitask}. These approaches leverage tensor structure to reduce model complexity while maintaining representational power. In the context of clustering, tensor-based methods have shown particular promise for multi-view and multi-way data analysis. Cichocki et al. (2015) \cite{cichocki2015tensor} developed tensor-based multiway arrays for big data sets, while Benjamin et al. (2024)  \cite{benjamin2024tensor} proposed tensorized possibilistic c-means for high-dimensional clustering. However, these methods are not designed for federated environments and lack the privacy-preserving mechanisms required for distributed deployment. 

\section{Mathematical Preliminaries}
\label{sec:Mathematical_Preliminaries}

This section presents a thorough mathematical formulation of the fundamental concepts and notations employed in this study. This includes an mathematical overview of fuzzy c-mean clustering, multi-view clustering, kernel Euclidean distance, tensor algebra, and tensor decomposition techniques, specifically PARAFAC2 and Tucker decomposition. These mathematical preliminaries provide the necessary foundation for understanding the proposed federated multi-view clustering framework.
\subsection{Multi-View Clustering}

\textbf{Notation and Mathematical Framework:} Let $X = \{x_1,\ldots,x_n\}$ be the collection of $n$ data points in $\mathbb{R}^d$ with $x_i=[x_{ij}]_{n \times d}$, $i=1,\ldots,n$, $j=1,\ldots,d$; $A = \{a_1,\ldots,a_c\}$ be the set of cluster centers with $a_k=[a_{kj}]_{c\times d}$ and $k=1,\ldots,c$, and $U = \{\mu_1,\ldots,\mu_c\}$ be the cluster membership matrix with $\mu_k=[\mu_{ik}]_{n\times c}$. 

In the multi-view scenario, $h$ denotes the view index, $h = 1, \ldots, s$, and the feature dimensionality of the $h$-th view is denoted as $d_h$. Thus, $X^h= \{x_1^h,\ldots,x_n^h\}$ defines the data input $X$ in the $h$-th view with $x_i^h=[x_{ij}^h]_{n\times d_h}$. Hence, $A^h = \{a_1^h,\ldots,a_c^h\}_{h=1}^s$ represents the cluster centers of the $h$-th view with $a_k^h=[a_{kj}^h]_{c \times d_h}$. 

\textbf{Soft Clustering:} The mathematical foundations of multi-view clustering build upon single-view clustering formulations. In this study, we focus on soft clustering approaches that provide probabilistic cluster assignments, enabling better uncertainty quantification in federated environments. The objective function of standard fuzzy c-means (FCM) or soft clustering can be expressed as:

\begin{equation}
J_{FCM}(U,A) = \sum_{i=1}^n \sum_{k=1}^c \mu_{ik}^m \|x_i -a_k\|^2 
\label{eqn:standardFCM}
\end{equation}

where $U$ is the membership matrix of data sets $X$ in clusters $c$, $A$ is the cluster centers matrices of the $j$-th feature in clusters $c$, and $m$ is the fuzzifier to control the distribution of membership matrix $U$. Eq. \ref{eqn:standardFCM} is designed to address data with a single representation, which is not compatible to solve a problem with multiple representations. To tackle this, $s$ number of views is introduced, such as indicator $h= 1, \ldots, s$ enabled the representation of data X from $s$ views. In this way, the clustering problems based on multiple representation data $X^h$ can be solved by using the following equation:

\begin{equation}
J_{MVC} (U^h, A^h) = \sum_{h=1}^s \sum_{i=1}^n \sum_{k=1}^c (\mu_{ik}^h)^{m} \|x_i^h - a_k^h\|^2
\label{eqn:MVC}
\end{equation}

As displayed in Eq. \ref{eqn:MVC}, the memberships of the $h$-th view data in $c$ clusters $\mu_{ik}^h$ is computed locally. Usually, to get a global memberships of data view $s$, we have an additional step called as aggregation of collected memberships across data views $s$. There are multiple ways to aggregate the memberships of $s$ data views, i.e., using the averagation approach or taking a product of memberships of the $i$-th data point across all views $s$ in clusters $c$ and divide it with the number of data points. If we consider a weight factor $v_h$, we can multiply each weight factors with the memberships and divide it by the number of data views, etc. As we focused on a client' multi-view data, we simplify the computation for each view memberships by directly implementing the global memberships $U^*$ from an early iteration. In this way, we can modify Eq. \ref{eqn:MVC} in the following way

 \begin{equation}
J_{{MVC}_2} (U^*, A^h) = \sum_{h=1}^s \sum_{i=1}^n \sum_{k=1}^c (\mu_{ik}^*)^m \|x_i^h - a_k^h\|^2
\label{eqn:MVCv2}
\end{equation}

The objective is to provide a comprehensive mathematical objective function that can effectively handle multi-view data distributed across multiple clients using tensor algebra, taking into account non-linearity compatibilities. Consequently, the distance between data points and cluster centers across data views in Euclidean space is replaced by a kernel Euclidean distance (KED). The following definition is proposed for the mathematical expression of KED:

\begin{equation}
\text{KED} (x_{ij}^h, a_{kj}^h) = \bigg \{ 1 - \text{exp} \bigg( - \sum_{j=1}^{d_h} \delta_{ij}^h (x_{ij}^h - a_{kj}^h)^2 \bigg ) \bigg \}
\label{eqn:KED}
\end{equation}

where $ \delta_{ij}^h$ is a heat kernel coefficient (HKC) to control the Euclidean distance between data points $x_{ij}^h$ and cluster centers $a_{kj}^h$ across data views in kernel space. By substituting the Euclidean distance in Eq. \ref{eqn:MVCv2} with the KED defined in Eq. \ref{eqn:KED}, we obtain the following objective function for alternative multi-view clustering (AMVC):

 \begin{equation}
J_{AMVC} (U^*, A^h) = \sum_{h=1}^s \sum_{i=1}^n \sum_{k=1}^c (\mu_{ik}^*)^m \text{KED} (x_{ij}^h, a_{kj}^h) 
\label{eqn:MVCv3}
\end{equation}

subject to:
\begin{equation*}
\sum_{k=1}^c \mu_{ik}^* = 1, \mu_{ik}^* \in [0, 1]
\end{equation*}

As illustrated in Eq. \ref{eqn:MVCv3}, it is imperative to prioritize all data views equally. This approach is not conducive to the efficient resolution of multiple resources data. Given that divergent perspectives correspond to disparate manifestations of feature behavior, it becomes imperative to allocate a distinct weight, denoted as $v_h$, to each data view, denoted as $x_i^h$. It is imperative to underscore that this allocation must be completed, in accordance with the principle that  $\sum_{h=1}^s v_h = 1$. By incorporating the weight factors, denoted by $v_h$, into the analysis, a modification to the Eq. \ref{eqn:MVCv3} is proposed, resulting in the following objective function for the efficient kernel multi-view clustering (E-KMVC):

\begin{equation}
J_{E-KMVC} (V_h, U^*, A^h) = \sum_{h=1}^s v_h^\alpha \sum_{i=1}^n \sum_{k=1}^c (\mu_{ik}^*)^m \bigg \{ 1 - \text{exp} \big( - \sum_{j=1}^{d_h} \delta_{ij}^h (x_{ij}^h - a_{kj}^h)^2 \big ) \bigg \}
\label{eqn:E-KMVC}
\end{equation}

subject to:
\begin{equation*}
\sum_{k=1}^c \mu_{ik}^* = 1, \mu_{ik}^* \in [0, 1], \quad \sum_{h=1}^s v_h = 1, v_h \in [0, 1]
\end{equation*}

where $\alpha$ is an exponent parameter to control the distribution of data views during the clustering processes. Eq. \ref{eqn:E-KMVC} serves as the foundational objective function for the proposed efficient federated kernel multi-view clustering (E-FKMVC) framework. This framework is designed to address the challenges associated with multi-view data clustering in federated learning environments, particularly focusing on data heterogeneity, privacy preservation, and communication efficiency.

\subsection{Tensor Decomposition Methods}

\textbf{Tensor Notations:} In this subsection, we introduce some basic notations and definitions of tensor algebra. Most of the overviews and notations are borrowed from Kolda et al. (2009) \cite{kolda2009tensor}.

\begin{figure}[!htbp]
\centering
\includegraphics[width=0.6\textwidth]{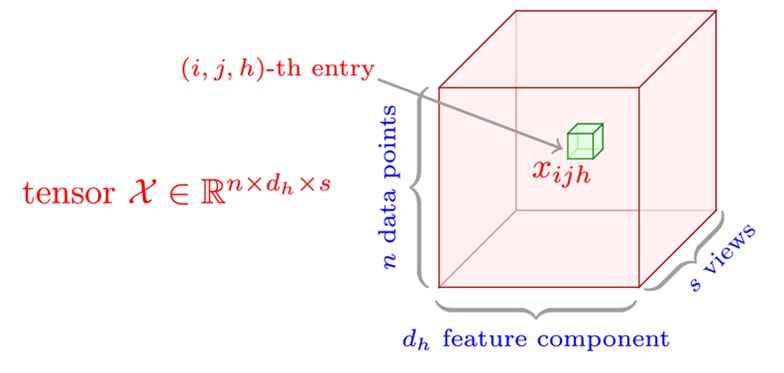}
\caption{Tensor Representation of Multi-View Data with Different Modalities. The number of data points $n$ is represented along the first mode, the feature dimensionality $d_h$ of each view is represented along the second mode, and the number of views $s$ is represented along the third mode. Each slice along the third dimension corresponds to a different view of the same data points, capturing complementary information across modalities.}
\label{fig:MVDatainTensorSpace}
\end{figure}

A tensor is a multi-dimensional array that generalizes the concepts of scalars (0th-order tensors), vectors (1st-order tensors), and matrices (2nd-order tensors) to higher dimensions. An $N$-way or $N$-th order tensor is denoted by a calligraphic letter, e.g., $\mathcal{X} \in \mathbb{R}^{I_1 \times I_2 \times \ldots \times I_N}$, where $I_n$ represents the size of the tensor along the $n$-th mode. The elements of the tensor are accessed using multiple indices, e.g., $\mathcal{X}_{i_1,i_2,\ldots,i_N}$. Consider a multi-view data $\mathcal{X}=\{x_{ijh}\}_{i=1,j=1,h=1}^{n, d_h, s}$ or equivalently as a set of 2D matrices $\mathcal{X}=\{X^1,\ldots,X^s\}$ with size $n \times d_h$, where $i$ indicates the data point index, $j$ indicates the feature index, and $h$ indicates the view index. Each slice along the third mode represents data from a specific view (see Fig. \ref{fig:MVDatainTensorSpace}).

\subsubsection{The Basic Background of Tensors}

This subsection provides some definitions and concepts of tensor algebra, most of overviews and notations are borrowed from Kolda et al. \cite{kolda2009tensor}. 
 
Assume that $\mathcal{A} \in \mathbb{R}^{G_1\times G_2\times \ldots \times G_n \times \ldots \times G_N} $ is an $N$-th order tensor, where $G_n$ represents the size of the tensor along the $n$-th mode. The elements of the tensor are accessed using multiple indices, e.g., $\mathcal{A}_{g_1,g_2,\ldots,g_n}$.

\textit{Definition 1:} The $N$-th order of tensor $\mathcal{A}$ is expressed as 
\begin{equation}
\mathcal{A}=[\mathcal{A}_{g_1,g_2,\ldots,g_n } ]\in \mathbb{R}^{G_1\times G_2\times \ldots \times G_n \times \ldots \times G_N}  
\end{equation}

where the $(g_1,g_2,\ldots,g_n )$-th elements of the tensor $\mathcal{A} \in \mathbb{R}^{G_1\times G_2\times \ldots \times G_n \times \ldots \times G_N} $ is containing $(\mathcal{A})_{g_1,g_2,\ldots,g_n}$ or $a_{g_1,g_2,\ldots,g_n}$ for indices $g_1=\{1,2,\ldots,G_1\}$, $g_2=\{1,2,\ldots,G_2\}$, and $g_n=\{1,2,\ldots,G_N\}$.

\begin{figure}[!htbp]
\centering
\includegraphics[width=0.5\textwidth]{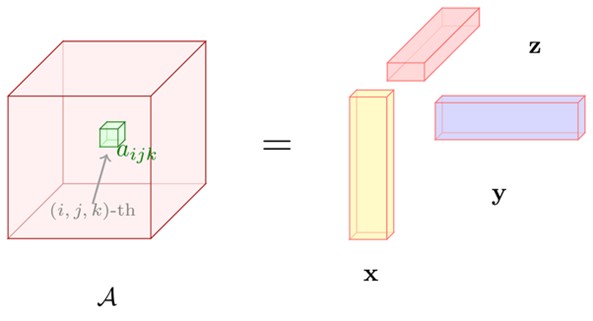}
\caption{The Illustration of Rank-One Third-Order Tensor $\mathcal{A} = x \circ y \circ z$. The $(i,j,k)$ element of $\mathcal{A}$ is given by $a_{ijk}=x_i y_j z_k$. In terms of element wise, the 3-way tensor can be written as $a_{ijk} \approx \sum_{r=1}^R x_{ir} y_{jr} z_{kr}$.}
\label{fig:rankonethirdordertensor}
\end{figure}

\textit{Definition 2:} The \textit{outer product} of $N$ \textit{vectors} of rank-one  $\mathcal{A} \in \mathbb{R}^{G_1\times G_2\times \ldots \times G_n \times \ldots \times G_N} $ is defined as

\begin{equation}
\mathcal{A}= a^{(1)} \circ a^{(2)} \circ \ldots \circ a^{(N)} 
 \end{equation}

where “$\circ$” represents the vector outer product. Fig. \ref{fig:rankonethirdordertensor} presented the illustration of rank-one third-order tensor $\mathcal{A} = x \circ y \circ z$. The $(i,j,k)$ element of $\mathcal{A}$ is given by $a_{ijk}=x_i y_j z_k$. In terms of element wise, the 3-way tensor can be written as $a_{ijk} \approx \sum_{r=1}^R x_{ir} y_{jr} z_{kr}$.

\textit{Definition 3:} The $n$-mode matrix product of   $\mathcal{A} \in \mathbb{R}^{G_1\times G_2\times \ldots \times G_n \times \ldots \times G_N} $ is denoted by  $\mathcal{A}_{(n)}\in \mathbb{R}^{G_1\times G_2\times \ldots \times G_{n-1} \times G_n \times G_{n+1} \times \ldots \times G_N} $  contains the element of $a_{g_1,g_2,\ldots,g_n}$ at  $g_n$-th row and at $j$th column, denoted as
\begin{equation}
 j=1+\sum_{\ell=1,\ell \neq n}^N (g_{\ell}-1) \prod_{m=1,m\neq n}^{\ell-1} G_m                 
 \end{equation}

\textit{Definition 4:} The $n$-\textit{mode product} between tensor $\mathcal{A}\in \mathbb{R}^{G_1 \times G_2 \times \ldots \times G_n \times \ldots \times G_N}$ and a matrix $L\in \mathbb{R}^{J_n \times G_n}$ is defined as $\mathcal{B}=\mathcal{A} \times_n L$, where the product of tensor $\mathcal{B}\in \mathbb{R}^{G_1 \times G_2 \times \ldots \times G_{n-1} \times G_n \times G_{n+1} \times \ldots \times G_N}$. In terms of element wise, the $n$-mode product of tensor $\mathcal{B}$ can be computed as 

\begin{equation}
b_{g_1g_2 \ldots g_{n-1} j_n g_{n+1} \ldots g_N} = \sum_{g_1 =1}^{g_n} a_{g_1 g_2 \ldots g_n \ldots g_N} b_{j_n g_n}
\end{equation}

\textit{Definition 5:} The \textit{tensor inner product} between tensor $\mathcal{A} \in \mathbb{R}^{G_1 \times G_2 \times \ldots \times G_n \times \ldots \times G_N}$ and $\mathcal{B} \in \mathbb{R}^{J_1 \times J_2 \times \ldots \times J_m \times \ldots  \times J_M}$ is denoted as

\begin{equation}
\mathcal{C}=〈\mathcal{A},\mathcal{B}〉_{n,m}
\end{equation}

The element wise of tensor inner product $\mathcal{A}$ and $\mathcal{B}$ can be written explicitly as below

\begin{equation}
c_{g_1g_2 \ldots g_{n-1} g_{n+1} \ldots g_N j_1j_2 \ldots j_{m-1} j_{m+1} \ldots j_M} = \sum_{q_1 =1}^{q_p} a_{g_1g_2 \ldots g_{n-1} g_q g_{n+1} \ldots g_N} b_{j_1j_2 \ldots j_{m-1} j_q j_{m+1} \ldots j_M}
\end{equation}
where $i_n = j_m = q_p$ and the tensor $\mathcal{C} \in \mathbb{R}^{G_1 \times \ldots \times G_{n-1} \times G_{n+1} \times\ldots \times G_N \times J_1 \times J_2 \times \ldots \times J_{m-1} \times J_{m+1} \times \ldots \times J_M}$.

\textit{Definition 6:} The \textit{tensor scalar product} of tensors  $\mathcal{A},\mathcal{B} \in  \coprod^{G_1 \times G_2 \times \ldots \times G_{n-1} \times G_{n+1} \times\ldots \times G_N}$ is the inner product of tensor $\mathcal{A}$ and $\mathcal{B}$, denoted as below

\begin{equation}
\langle {\mathcal{A}, \mathcal{B}} \rangle \buildrel \Delta \over = \sum_{g_1 =1}^{G_1} \sum_{g_2 = 1}^{G_2} \ldots \sum_{g_n =1}^{G_N} a_{g_1 g_2 \ldots g_{n-1} g_{n+1} \ldots g_N} b_{g_1 g_2 \ldots g_{n-1} g_{n+1} \ldots g_N}
\end{equation}

\textit{Definition 7:} The \textit{Frobenius norm} of tensor $\mathcal{A} \in \mathbb{R}^{G_1 \times G_2 \times \ldots \times G_n \times \ldots  \times G_N}$ also known as the “energy” of the tensor, defined as

\begin{equation}
\|\mathcal{A}\|_F = \varepsilon_{\mathcal{A}} \buildrel \Delta \over = \sqrt{ \langle {\mathcal{A}, \mathcal{A}} \rangle }
\end{equation}

\subsubsection{The Higher Order Singular Value Decomposition (HOSVD)}

Tensor factorization is a method for reducing a large volume of data to a smaller set of data points. This is achieved by leveraging the collinearity of its heterogeneous data \cite{wang2017tensor}. CANDECOMP/PARAFAC is a technique for decomposing data into its constituent parts. It is one of the most widely used tensor decomposition techniques, as referenced in the works of Kolda et al. (2009) \cite{kolda2009tensor}. and Kiers (2000) \cite{kiers2000towards}.  Tucker decomposition is another popular tensor decomposition technique that decomposes a tensor into a core tensor and factor matrices. It was first introduced by Tucker (1966) \cite{tucker1966some} and has been widely used in various applications. Higher-Order Singular Value Decomposition (HOSVD) is a generalization of the matrix singular value decomposition (SVD) to higher-order tensors. It was introduced by De Lathauwer et al. (2000) \cite{de2000multilinear} and has been widely used in various applications, including image processing, signal processing, and machine learning. In order to provide a comprehensive understanding of tensor decomposition techniques, we present the following three widely used methods:

The PARAFAC decomposition \cite{carroll1970analysis} of a tensor $\mathcal{A} \in \mathbb{R}^{G_1 \times G_2 \times \ldots \times G_N}$ is defined as 
\begin{equation}
\mathcal{A} \approx \sum_{r=1}^R \lambda_r a_r^{(1)} \circ a_r^{(2)} \circ \ldots \circ a_r^{(N)} 
\label{eqn:parafac}
\end{equation}
where $\lambda_r$ is the weight of the $r$-th rank-one tensor, and $a_r^{(n)} \in \mathbb{R}^{G_n}$ is the factor vector along the $n$-th mode. The minimum value of $R$ that satisfies Eq. \ref{eqn:parafac} is defined as the rank of tensor $\mathcal{A}$. 

The Tucker decomposition \cite{tucker1966some} of a tensor $\mathcal{A} \in \mathbb{R}^{G_1 \times G_2 \times \ldots \times G_N}$ is defined as
\begin{equation}
\mathcal{A} \approx \mathcal{S} \times_1 U^{(1)} \times_2 U^{(2)} \times_3 \ldots \times_N U^{(N)} 
\label{eqn:tucker}
\end{equation}
where $\mathcal{S} \in \mathbb{R}^{R_1 \times R_2 \times \ldots \times R_N}$ is the core tensor, and $U^{(n)} \in \mathbb{R}^{G_n \times R_n}$ is the factor matrix along the $n$-th mode. The dimensions $R_n$ are the multilinear ranks of tensor $\mathcal{A}$.

The Higher-Order Singular Value Decomposition (HOSVD) \cite{de2000multilinear} of a tensor $\mathcal{A} \in \mathbb{R}^{G_1 \times G_2 \times \ldots \times G_N}$ is defined as
\begin{equation}
\mathcal{A} = \mathcal{S} \times_1 U^{(1)} \times_2 U^{(2)} \times_3 \ldots \times_N U^{(N)} 
\label{eqn:hosvd}
\end{equation}
where $\mathcal{S} \in \mathbb{R}^{G_1 \times G_2 \times \ldots \times G_N}$ is the core tensor, and $U^{(n)} \in \mathbb{R}^{G_n \times G_n}$ is the orthogonal factor matrix along the $n$-th mode. The HOSVD provides a unique decomposition of tensor $\mathcal{A}$, and the factor matrices $U^{(n)}$ are obtained by performing SVD on the mode-$n$ unfoldings of tensor $\mathcal{A}$.

Table \ref{tab:tensordecomposition} summarizes the key characteristics of the three tensor decomposition methods discussed above.
\begin{table}[!htbp]
\centering
\caption{Summary of Tensor Decomposition Methods: Characteristics, Components, and Applications}
\begin{adjustbox}{width=1\columnwidth}
\begin{tabular}{|p{2cm}|p{4.5cm}|p{3cm}|p{4cm}|p{3.5cm}|}
\hline
\textbf{Method} & \textbf{Decomposition Form} & \textbf{Components} & \textbf{Key Features} & \textbf{Applications} \\
\hline
PARAFAC (CP) & $\mathcal{A} \approx \sum_{r=1}^R \lambda_r a_r^{(1)} \circ a_r^{(2)} \circ \ldots \circ a_r^{(N)}$ & 
\begin{itemize}[leftmargin=*, nosep]
    \item Weights $\lambda_r$
    \item Factor vectors $a_r^{(n)}$
\end{itemize} & 
\begin{itemize}[leftmargin=*, nosep]
    \item Unique under mild conditions
    \item Fixed rank $R$
    \item Interpretable factors
\end{itemize} & 
\begin{itemize}[leftmargin=*, nosep]
    \item Chemometrics
    \item Signal processing
    \item Neuroscience
\end{itemize} \\
\hline
Tucker & $\mathcal{A} \approx \mathcal{S} \times_1 U^{(1)} \times_2 U^{(2)} \times_3 \ldots \times_N U^{(N)}$ & 
\begin{itemize}[leftmargin=*, nosep]
    \item Core tensor $\mathcal{S}$
    \item Factor matrices $U^{(n)}$
\end{itemize} & 
\begin{itemize}[leftmargin=*, nosep]
    \item Flexible multilinear ranks $(R_1, R_2, \ldots, R_N)$
    \item Non-unique decomposition
    \item Higher compression ratio
\end{itemize} & 
\begin{itemize}[leftmargin=*, nosep]
    \item Image compression
    \item Dimensionality reduction
    \item Data mining
\end{itemize} \\
\hline
HOSVD & $\mathcal{A} = \mathcal{S} \times_1 U^{(1)} \times_2 U^{(2)} \times_3 \ldots \times_N U^{(N)}$ & 
\begin{itemize}[leftmargin=*, nosep]
    \item All-orthogonal core $\mathcal{S}$
    \item Orthonormal matrices $U^{(n)}$
\end{itemize} & 
\begin{itemize}[leftmargin=*, nosep]
    \item Unique decomposition
    \item Orthogonal factors
    \item Optimal truncation properties
\end{itemize} & 
\begin{itemize}[leftmargin=*, nosep]
    \item Multilinear PCA
    \item Subspace analysis
    \item Tensor initialization
\end{itemize} \\
\hline
\multicolumn{5}{|c|}{\textbf{Complexity Comparison}} \\
\hline
\textbf{Method} & \multicolumn{2}{c|}{\textbf{Time Complexity}} & \multicolumn{2}{c|}{\textbf{Space Complexity}} \\
\hline
PARAFAC (CP) & \multicolumn{2}{c|}{$\mathcal{O}(R \prod_{n=1}^N G_n)$ per iteration} & \multicolumn{2}{c|}{$\mathcal{O}(R \sum_{n=1}^N G_n)$} \\
\hline
Tucker & \multicolumn{2}{c|}{$\mathcal{O}(\prod_{n=1}^N R_n + \sum_{n=1}^N G_n R_n)$} & \multicolumn{2}{c|}{$\mathcal{O}(\prod_{n=1}^N R_n + \sum_{n=1}^N G_n R_n)$} \\
\hline
HOSVD & \multicolumn{2}{c|}{$\mathcal{O}(\sum_{n=1}^N G_n^2 \prod_{m \neq n} G_m)$} & \multicolumn{2}{c|}{$\mathcal{O}(\prod_{n=1}^N G_n + \sum_{n=1}^N G_n^2)$} \\
\hline
\end{tabular}
\end{adjustbox}
\label{tab:tensordecomposition}
\end{table}

\subsubsection{Tensor Distance}
In this subsection, we introduce the concept of tensor distance, which extends the traditional Euclidean distance to higher-order tensors. This is particularly relevant in the context of multi-view data represented as tensors, where capturing the relationships between different modes is crucial. 

The traditional Euclidean distance between two vectors $x_i$ and $a_k$ in $\mathbb{R}^d$ is defined as
\begin{equation}
d(x_i, a_k) = \|x_i -a_k\|^2
\end{equation}

where $\|\cdot\|$ denotes the Euclidean norm. However, when dealing with multi-view data represented as tensors, it is essential to consider the interactions across different modes. To address this, we propose the following definition of tensor distance:
 
\textit{Definition 8:} The \textit{tensor distance} of a given $\mathcal{X},\mathcal{A}\in \mathbb{R}^{G_1 \times G_2 \times \ldots  \times G_N}$  $(N\geq1)$, and $x$, $a$ denoted as the vector representation of tensor $\mathcal{X},\mathcal{A}$, such as the element wise of tensor $\mathcal{XA}_{G_1 \times G_2 \times \ldots \times G_N}$ $(1\leq g_i\leq G_n,1\leq i\leq N)$ in $\mathcal{X},\mathcal{A}$ is corresponding to $x_l$, $a_l$, i.e., the $l$-th element in $x$, the $l$-th element in $a$, where $l=1+\sum_{i'=1,i'\neq i}^{G_1 \times G_2 \times \ldots \times G_N} (g_l-1) \prod_{m=1,m\neq i}^{i'-1} G_m$. Then the tensor distance (TD) between $\mathcal{X}$ and $\mathcal{A}$  can be defined as

\begin{equation}
d_{TD} (\mathcal{X}, \mathcal{A}) = \sqrt{ \sum_{l,m = 1}^{G_1 \times G_2 \times \ldots  \times G_N} \rho_{lm} (x_l - a_l) (x_m - a_m)} = \sqrt{ (x-a)^T P (x-a) }
\label{eqn:tensordistance}
\end{equation}

In Eq. \ref{eqn:tensordistance}, the tensor distance $d_{TD} (\mathcal{X}, \mathcal{A})$ is computed using a weighted Euclidean distance formulation, formulated in terms of the vectorized representations $x$ and $a$ of the tensors $\mathcal{X}$ and $\mathcal{A}$, respectively. The matrix $P$ is a metric matrix that incorporates the heat-kernel coefficients $\rho_{lm}$, which modulate the contribution of each pair of elements $(x_l, a_l)$ and $(x_m, a_m)$ to the overall distance. The metric coefficients $\rho_{lm}$ are defined as follows:

\begin{equation}
\rho_{lm} = \frac{1}{2\pi \theta} \frac{\text{ exp } \big\{ \|q_1 - q_m\|_2^2\big \}} {2\theta} 
\label{eqn:heatkernelcoefficients}
\end{equation}

In Eq. \ref{eqn:heatkernelcoefficients}, the term $\|q_1 - q_m\|_2$ represents the Euclidean distance between two points $q_1$ and $q_m$ in the feature space, which can be computed as:

\begin{equation}
\|q_1 - q_m\|_2 = \sqrt{ (g_1 +g_1')^2  + (g_2 +g_2')^2 +\ldots + (g_n +g_n')^2 +\ldots + (g_N +g_N')^2} 
\label{eqn:euclideandistance}
\end{equation}

In Eq. \ref{eqn:euclideandistance}, $g_n$ and $g_n'$ denote the indices along the $n$-th mode of tensors $\mathcal{X}$ and $\mathcal{A}$, respectively. The parameter $\theta$ in Eq. \ref{eqn:heatkernelcoefficients} is a scaling factor that controls the spread of the heat kernel, influencing how rapidly the weights decay with increasing distance between points in the feature space.

\subsection{Summary and Research Gaps}

A review of extant literature reveals several significant research gaps that motivate the proposed approach, including the limited integration of research findings. While federated learning, multi-view clustering, and tensor decomposition have been the focus of extensive research study, their integration remains largely unexplored. Existing federated clustering methods do not leverage multi-view structure, and multi-view clustering approaches are not designed for federated environments. In the context of communication inefficiency, contemporary federated multi-view methods do not leverage tensor structure for the purpose of communication compression. The high-dimensional nature of multi-view data exacerbates communication overhead in federated settings, requiring novel tensor-aware compression schemes.

Existing privacy-preserving mechanisms are not optimized for multi-view tensor data, potentially leading to information leakage across views or reduced clustering utility. The development of novel differential privacy mechanisms tailored to tensor structures is imperative. A further issue is the absence of personalization. The prevailing assumption underlying most federated clustering approaches is that data distributions across clients are homogeneous. In practice, clients frequently possess heterogeneous data characteristics, necessitating customized clustering solutions that adapt to local data patterns while leveraging global knowledge. As indicated in the theoretical understanding, the convergence properties and privacy guarantees of federated multi-view tensor clustering remain to be fully elucidated. A theoretical analysis is necessary to provide convergence guarantees and characterize the privacy-utility trade-offs.

The proposed frameworks address these limitations by integrating advanced tensor decomposition with personalized federated learning, leveraging a heat-kernel-enhanced tensorized approach. The aforementioned framework provides both theoretical guarantees and practical advantages.

\section{Proposed Methodology}
\label{sec:proposed_methodology}

This section presents the proposed federated heat-kernel enhanced multi-view clustering framework, which does not involve the sharing of raw data across clients. The section commences with the formulation of the problem and the introduction of the federated kernel Euclidean distance (FKED) metric. Subsequently, the federated heat-kernel coefficients (FedH-KC) are defined, which adaptively weight feature distances in the federated setting. In summary, the overarching objective function for federated multi-view clustering using FKED and FedH-KC is hereby presented.

\subsection{Problem Formulation}

Let $\mathcal{L} = \{L_1, L_2, \ldots, L_M\}$ be a set of $M$ clients in the federated system. Each client $L_{\ell}$ possesses local multi-view data that can be organized into tensor form $\mathcal{X}_{[\ell]} \in \mathbb{R}^{n(\ell) \times d_1^{(\ell)} \times d_2^{(\ell)} \times \cdots \times d_{s(\ell)}^{(\ell)}}$, where:
\begin{itemize}
\item $n(\ell)$ is the number of samples at client $\ell$
\item $s(\ell)$ is the number of views at client $\ell$ 
\item $d_h^{(\ell)}$ is the feature dimension of view $h$ at client $\ell$
\end{itemize}

Alternatively, we can represent the data as a collection of view-specific matrices $X_{[\ell]} = \{X_{[\ell]}^h\}_{h=1}^{s(\ell)}$ where $X_{[\ell]}^h \in \mathbb{R}^{n(\ell) \times d_h^{(\ell)}}$, which can be tensorized for advanced decomposition techniques.

\subsection{Federated Kernel Euclidean Distance}

In federated environment, we assume participated clients shared different number of data but the same feature component. Thus, the ground truth or predicted clusters of multi-view data sets stored on their data centers may vary. In this way, each data points on clients will shape a unified global multi-view datasets in a certain problem or case. Therefore, it is explained a lot the definition of \textit{``heterogeneous''} on data. This characteristic is denoted by $c(\ell)$, representing the number of clusters of multi-view data sets held by the $\ell$-th client. By following this direction, we extend the heat-kernel enhanced distance metric KED in Eq. \ref{eqn:KED} to operate in federated environment. For client $\ell$, the kernel Euclidean distance (KED) between $x_{[\ell]ij}^h$ (representing data point $i$, $\forall i = 1, \ldots,n (\ell)$) and cluster center tensor $a_{[\ell]kj}^h$ (representing cluster $k$, $\forall k = 1, \ldots, c(\ell)$) is:

\begin{equation}
\text{FKED}( x_{[\ell]ij}^h, a_{[\ell]kj}^h ) = 1 - \exp \left( - \sum_{h=1}^{s(\ell)} \sum_{j=1}^{d_h^{(\ell)}} \delta_{[\ell]i j}^{h} \left(x_{[\ell]ij}^h - a_{[\ell]kj}^h \right)^2 \right)
\label{eqn:federated_ked}
\end{equation}

where $\delta_{[\ell]i j}^h$ are heat-kernel coefficients held by $M$ clients. 

\subsection{Federated Heat-Kernel Coefficients}

We focus on developing federated heat-kernel coefficients (FedH-KC), which are computed locally by each client to transform feature distances into privacy-preserving exponential kernel measures tailored to the federated setting. For the $j$-th feature in view $h$, we define FedH-KC $\delta _{\left[ \ell  \right]ij}^h$ using two distinct estimators.

\begin{equation}
\label{eqn:fed_delta}
\text{FedH-KC} = \delta _{\left[ \ell  \right]ij}^h = \frac{
    x_{\left[ \ell  \right]ij}^h - \min_{1 \le i \le n(\ell)} \left( x_{\left[ \ell  \right]ij}^h \right)
}{
    \max_{1 \le i \le n(\ell)} \left( x_{\left[ \ell  \right]ij}^h \right) - \min_{1 \le i \le n(\ell)} \left( x_{\left[ \ell  \right]ij}^h \right) + \epsilon
}
\qquad \forall j, h, \ell
\end{equation}

where $\epsilon > 0$ prevents division by zero when features have identical values. This normalizes features within each view for consistent distance calculations across clients with different data distributions. and

\begin{equation}
\text{FedH-KC} = \delta _{\left[ \ell  \right]ij}^h = \left| {x_{\left[ \ell  \right]ij}^h - {{\bar x}_{\left[ \ell  \right]j}}^h} \right| \qquad \forall j, h, \ell
\label{eqn:fed_delta_v2}
\end{equation}
where $\bar{x}_{[\ell]j}^h = \frac{1}{n(\ell)}\sum_{i = 1}^{n(\ell)} x_{[\ell]ij}^h$ represents the mean value of the $j$-th feature in view $h$ for client $\ell$, harmonizing with the centralized case notation.

\subsection{Personalized Federated Learning}

Considers there are $M$ number of clients storing multi-view data sets $X_{[\ell]} =X_{[\ell]i}^h$ for all $\ell=1, \ldots, M$ and $h=1,\ldots,s(\ell)$.  Here, we assume each client is storing a different number of dimensionality of data points $x_i$. To solve clustering problem for these multi-view data sets distributed on $M$ clients, we propose an efficient federated kernel multi-view clustering (E-FKMVC). The proposed E-FKMVC modified Eq. \ref{eqn:E-KMVC} in the following way

\begin{equation}
\begin{adjustbox}{width=0.9\columnwidth}
$ 
J_{E-FKMVC}^{(\ell)} = J_{E-FKMVC} (V_{[\ell]}, U_{[\ell]}^*, A_{[\ell]}^h) = \sum_{\ell=1}^M \sum_{h=1}^{s(\ell)} v_{[\ell]h}^\alpha \sum_{i=1}^{n(\ell)} \sum_{k=1}^{c(\ell)} (\mu_{[\ell]ik}^*)^m \text{~FKED}( x_{[\ell]ij}^h, a_{[\ell]kj}^h ) 
$
\end{adjustbox}
\label{eqn:E-FKMVC}
\end{equation}

subject to:
\begin{equation}
    \sum_{k=1}^c \mu_{[\ell]ik}^* = 1, \mu_{[\ell]ik}^* \in [0, 1]  \label{eqn:fed_constraint1}
\end{equation}
\begin{equation}
    \sum_{h=1}^s v_{[\ell]h} = 1, v_{[\ell]h} \in [0, 1] \label{eqn:fed_constraint2}
\end{equation}

Due to the non-convex nature of the E-FKMVC objective function in Eq. \ref{eqn:E-FKMVC}, direct minimization is not feasible. We employ an alternating optimization strategy that iteratively updates the membership matrix $U_{[\ell]}^*$, view weights $V_{[\ell]}$, and cluster centers $A_{[\ell]}^h$ until convergence.

To derive the update rules, we formulate the Lagrangian incorporating the normalization constraints:

\begin{equation}
\begin{split}
\tilde J_{E-FKMVC}^{(\ell)} &= J_{E-FKMVC} (V_{[\ell]}, U_{[\ell]}^*, A_{[\ell]}^h) \\
&\quad + \sum_{i=1}^{n(\ell)} {\lambda_{1i}} \left( \sum_{k=1}^{c(\ell)} {\mu_{[\ell]ik}^* - 1} \right) + {\lambda_2}\left( \sum_{h=1}^{s(\ell)} {v_{[\ell]h} - 1} \right)
\end{split}
\label{eqn:LAG_EFKMVFC}
\end{equation}

where ${\lambda _{1i}}$ and ${\lambda _2}$ are Lagrange multipliers enforcing the fuzzy clustering constraints: membership normalization and view weight normalization, respectively.

\begin{Theorem}[E-FKMVC Update Rules]
\label{thm:efkmvc_update}

For the federated objective function $J_{E-FKMVC}$ defined in Eq. \ref{eqn:E-FKMVC}, the necessary conditions for optimality yield the following update rules for client $\ell$:

\textbf{Membership Matrix Update:}
\begin{equation}
    \mu _{\left[ \ell  \right]ik}^* = \frac{{{{\left( {\sum\limits_{h = 1}^{s\left( \ell  \right)} {v_{\left[ \ell  \right]h}^\alpha {\rm{FKED}}\left( {x_{\left[ \ell  \right]ij}^h,a_{\left[ \ell  \right]kj}^h} \right)} } \right)}^{ - {{\left( {m - 1} \right)}^{ - 1}}}}}}{{\sum\limits_{k' = 1}^{c\left( \ell  \right)} {{{\left( {\sum\limits_{h = 1}^{s\left( \ell  \right)} {v_{\left[ \ell  \right]h}^\alpha {\rm{FKED}}\left( {x_{\left[ \ell  \right]ij}^h,a_{\left[ \ell  \right]k'j}^h} \right)} } \right)}^{ - {{\left( {m - 1} \right)}^{ - 1}}}}} }}
    \label{eqn:diff_U_EFKMVFC}
\end{equation}

\textbf{Cluster Centers Update:}
\begin{equation}
    a_{\left[ \ell  \right]kj}^h = \frac{{\sum\limits_{i = 1}^{n\left( \ell  \right)} {{{\left( {\mu _{\left[ \ell  \right]ik}^*} \right)}^m}v_{\left[ \ell  \right]h}^\alpha \exp \left( { - \delta _{\left[ \ell  \right]ij}^h{{\left\| {x_{\left[ \ell  \right]i}^h - a_{\left[ \ell  \right]k}^h} \right\|}^2}} \right)} }}{{\sum\limits_{i = 1}^{n\left( \ell  \right)} {{{\left( {\mu _{\left[ \ell  \right]ik}^*} \right)}^m}v_{\left[ \ell  \right]h}^\alpha \exp \left( { - \delta _{\left[ \ell  \right]ij}^h{{\left\| {x_{\left[ \ell  \right]i}^h - a_{\left[ \ell  \right]k}^h} \right\|}^2}} \right)} }}x_{\left[ \ell  \right]ij}^h
    \label{eqn:diff_A_EFKMVFC}
\end{equation}

\textbf{View Weights Update:}
\begin{equation}
    {v_{\left[ \ell  \right]h}} = \frac{{{{\left( {\sum\limits_{i = 1}^{n\left( \ell  \right)} {\sum\limits_{k = 1}^{c\left( \ell  \right)} {{{\left( {\mu _{\left[ \ell  \right]ik}^*} \right)}^m}{\rm{FKED}}\left( {x_{\left[ \ell  \right]ij}^h,a_{\left[ \ell  \right]kj}^h} \right)} } } \right)}^{ - {{\left( {\alpha  - 1} \right)}^{ - 1}}}}}}{{\sum\limits_{h' = 1}^{s\left( \ell  \right)} {{{\left( {\sum\limits_{i = 1}^{n\left( \ell  \right)} {\sum\limits_{k = 1}^{c\left( \ell  \right)} {{{\left( {\mu _{\left[ \ell  \right]ik}^*} \right)}^m}{\rm{FKED}}\left( {x_{\left[ \ell  \right]ij}^{h'},a_{\left[ \ell  \right]kj}^{h'}} \right)} } } \right)}^{ - {{\left( {\alpha  - 1} \right)}^{ - 1}}}}} }}
    \label{eqn:diff_V_EFKMVFC}
\end{equation}

where the updates are performed iteratively until convergence, subject to the normalization constraints in Eqs. \ref{eqn:fed_constraint1} and \ref{eqn:fed_constraint2}.

\end{Theorem}

\begin{proof}
We establish the necessary optimality conditions for the E-FKMVC objective function using Lagrangian optimization and derive the closed-form update rules for each parameter set. The proof proceeds by applying the method of Lagrange multipliers to the constrained optimization problem and solving the resulting system of equations. This classical optimization technique allows us to find optimal solutions while respecting the normalization constraints on memberships $U_{[\ell]}^*$ and view weights $V_{[\ell]}$.

\textbf{Part I: Membership Matrix Update Rule Derivation}

The membership matrix $U_{[\ell]}^*$ determines how strongly each data point $i$ belongs to each cluster $k$. To derive the optimal update equation for membership coefficients $\mu_{[\ell]ik}^*$ for client $\ell$, we employ the method of Lagrange multipliers. This approach transforms our constrained optimization problem into an unconstrained one by incorporating the constraint $\sum_{k=1}^{c(\ell)} \mu_{[\ell]ik}^* = 1$ directly into the objective function.

We begin by computing the partial derivative of the Lagrangian in Eq. \ref{eqn:LAG_EFKMVFC} with respect to $\mu_{[\ell]ik}^*$. This derivative tells us how the objective function changes when we slightly adjust the membership of data point $i$ in cluster $k$:
\begin{align}
\frac{\partial \tilde{J}_{E-FKMVC}^{(\ell)}}{\partial \mu_{[\ell]ik}^*} &= \frac{\partial}{\partial \mu_{[\ell]ik}^*} \left[\sum_{h=1}^{s(\ell)} v_{[\ell]h}^\alpha (\mu_{[\ell]ik}^*)^m \text{FKED}(x_{[\ell]i}^h, a_{[\ell]k}^h)\right] + \lambda_{1i} \\
&= m(\mu_{[\ell]ik}^*)^{m-1} \sum_{h=1}^{s(\ell)} v_{[\ell]h}^\alpha \text{FKED}(x_{[\ell]i}^h, a_{[\ell]k}^h) + \lambda_{1i} \label{eqn:lagrange_u_fed}
\end{align}

Here, the power rule of differentiation yields the factor $m(\mu_{[\ell]ik}^*)^{m-1}$, while $\lambda_{1i}$ is the Lagrange multiplier that enforces the membership normalization constraint for data point $i$.

At the optimal solution, this derivative must equal zero (a necessary condition for optimality). Setting the derivative to zero gives us:
\begin{equation}
m(\mu_{[\ell]ik}^*)^{m-1} \sum_{h=1}^{s(\ell)} v_{[\ell]h}^\alpha \text{FKED}(x_{[\ell]i}^h, a_{[\ell]k}^h) + \lambda_{1i} = 0
\end{equation}

Rearranging this equation to isolate $\mu_{[\ell]ik}^*$, we first solve for $(\mu_{[\ell]ik}^*)^{m-1}$ and then take the $(m-1)$-th root:
\begin{align}
(\mu_{[\ell]ik}^*)^{m-1} &= -\frac{\lambda_{1i}}{m \sum_{h=1}^{s(\ell)} v_{[\ell]h}^\alpha \text{FKED}(x_{[\ell]i}^h, a_{[\ell]k}^h)} \\
\mu_{[\ell]ik}^* &= \left(-\frac{\lambda_{1i}}{m}\right)^{1/(m-1)} \left(\sum_{h=1}^{s(\ell)} v_{[\ell]h}^\alpha \text{FKED}(x_{[\ell]i}^h, a_{[\ell]k}^h)\right)^{-1/(m-1)} \label{eqn:u_intermediate_fed}
\end{align}

At this point, $\lambda_{1i}$ remains unknown. To determine its value, we apply the normalization constraint which requires that all membership values for data point $i$ across all clusters must sum to unity (i.e., $\sum_{k=1}^{c(\ell)} \mu_{[\ell]ik}^* = 1$). Summing Eq. \ref{eqn:u_intermediate_fed} over all clusters $k$:
\begin{align}
\sum_{k=1}^{c(\ell)} \mu_{[\ell]ik}^* &= \left(-\frac{\lambda_{1i}}{m}\right)^{1/(m-1)} \sum_{k=1}^{c(\ell)} \left(\sum_{h=1}^{s(\ell)} v_{[\ell]h}^\alpha \text{FKED}(x_{[\ell]i}^h, a_{[\ell]k}^h)\right)^{-1/(m-1)} = 1
\end{align}

This equation allows us to solve for the term containing the Lagrange multiplier:
\begin{align}
\left(-\frac{\lambda_{1i}}{m}\right)^{1/(m-1)} &= \frac{1}{\sum_{k'=1}^{c(\ell)} \left(\sum_{h=1}^{s(\ell)} v_{[\ell]h}^\alpha \text{FKED}(x_{[\ell]i}^h, a_{[\ell]k'}^h)\right)^{-1/(m-1)}} \label{eqn:lambda_solution_fed}
\end{align}

Substituting this expression back into Eq. \ref{eqn:u_intermediate_fed} eliminates the Lagrange multiplier and yields the final closed-form update rule:
\begin{equation}
\mu_{[\ell]ik}^* = \frac{\left(\sum_{h=1}^{s(\ell)} v_{[\ell]h}^\alpha \text{FKED}(x_{[\ell]i}^h, a_{[\ell]k}^h)\right)^{-1/(m-1)}}{\sum_{k'=1}^{c(\ell)} \left(\sum_{h=1}^{s(\ell)} v_{[\ell]h}^\alpha \text{FKED}(x_{[\ell]i}^h, a_{[\ell]k'}^h)\right)^{-1/(m-1)}}
\end{equation}

This formula has an intuitive interpretation: the membership of data point $i$ in cluster $k$ is inversely proportional to the weighted distance (across all views) between the data point and the cluster center. Data points closer to a cluster center receive higher membership values for that cluster. The negative exponent $-1/(m-1)$ ensures this inverse relationship, while the denominator normalizes the memberships to sum to one. This establishes the membership matrix update rule in Eq. \ref{eqn:diff_U_EFKMVFC}.

\textbf{Part II: Cluster Centers Update Rule Derivation}

The cluster centers $a_{[\ell]kj}^h$ represent the prototype or centroid of each cluster in each view. To find optimal cluster centers, we differentiate the objective function in Eq. \ref{eqn:E-FKMVC} with respect to $a_{[\ell]kj}^h$, which represents the $j$-th feature coordinate of the $k$-th cluster center in view $h$ held by client $\ell$:

\begin{align}
\frac{\partial J_{E-FKMVC}^{(\ell)}}{\partial a_{[\ell]kj}^h} &= \frac{\partial}{\partial a_{[\ell]kj}^h} \left[v_{[\ell]h}^\alpha \sum_{i=1}^{n(\ell)} (\mu_{[\ell]ik}^*)^m \text{FKED}(x_{[\ell]i}^h, a_{[\ell]k}^h)\right] \\
&= v_{[\ell]h}^\alpha \sum_{i=1}^{n(\ell)} (\mu_{[\ell]ik}^*)^m \frac{\partial \text{FKED}(x_{[\ell]i}^h, a_{[\ell]k}^h)}{\partial a_{[\ell]kj}^h} \label{eqn:cluster_derivative_fed}
\end{align}

The key challenge here is computing the derivative of the FKED function, which involves an exponential of a sum of squared differences. Recall the definition of FKED from Eq. \ref{eqn:federated_ked}:
\begin{equation}
\text{FKED}(x_{[\ell]i}^h, a_{[\ell]k}^h) = 1 - \exp\left(-\sum_{j'=1}^{d_{[\ell]}^h} \delta_{[\ell]ij'}^h (x_{[\ell]ij'}^h - a_{[\ell]kj'}^h)^2\right)
\end{equation}

To simplify the derivation, we introduce a shorthand notation for the argument of the exponential:
$$\phi_{[\ell]ik}^h = \sum_{j'=1}^{d_h^{(\ell)}} \delta_{[\ell]ij'}^h (x_{[\ell]ij'}^h - a_{[\ell]kj'}^h)^2$$

This $\phi_{[\ell]ik}^h$ represents the heat-kernel weighted squared distance between data point $i$ and cluster center $k$ in view $h$ held by client $\ell$. Using the chain rule of differentiation (which states that the derivative of a composite function equals the derivative of the outer function times the derivative of the inner function):
\begin{align}
\frac{\partial \text{FKED}}{\partial a_{[\ell]kj}^h} &= \frac{\partial}{\partial a_{[\ell]kj}^h} \left[1 - \exp(-\phi_{[\ell]ik}^h)\right] \\
&= \exp(-\phi_{[\ell]ik}^h) \frac{\partial \phi_{[\ell]ik}^h}{\partial a_{[\ell]kj}^h} \\
&= \exp(-\phi_{[\ell]ik}^h) \cdot 2\delta_{[\ell]ij}^h (x_{[\ell]ij}^h - a_{[\ell]kj}^h) \cdot (-1) \\
&= -2\delta_{[\ell]ij}^h (x_{[\ell]ij}^h - a_{[\ell]kj}^h) \exp\left(-\sum_{j'=1}^{d_h^{(\ell)}} \delta_{[\ell]ij'}^h (x_{[\ell]ij'}^h - a_{[\ell]kj'}^h)^2\right) \label{eqn:fked_derivative_fed}
\end{align}

The factor of 2 arises from differentiating the squared term, and the negative sign comes from differentiating $(x - a)^2$ with respect to $a$.

Substituting this result into Eq. \ref{eqn:cluster_derivative_fed} and setting the derivative equal to zero (the optimality condition):
\begin{align}
&v_{[\ell]h}^\alpha \sum_{i=1}^{n(\ell)} (\mu_{[\ell]ik}^*)^m \left[-2\delta_{[\ell]ij}^h (x_{[\ell]ij}^h - a_{[\ell]kj}^h) \exp\left(-\sum_{j'=1}^{d_h^{(\ell)}} \delta_{[\ell]ij'}^h (x_{[\ell]ij'}^h - a_{[\ell]kj'}^h)^2\right)\right] = 0
\end{align}

Since $v_{[\ell]h}^\alpha > 0$, $\delta_{[\ell]ij}^h > 0$, and the exponential term is always positive, we can simplify by dividing out these non-zero factors:
\begin{align}
\sum_{i=1}^{n(\ell)} (\mu_{[\ell]ik}^*)^m &\exp\left(-\sum_{j'=1}^{d_h^{(\ell)}} \delta_{[\ell]ij'}^h (x_{[\ell]ij'}^h - a_{[\ell]kj'}^h)^2\right) (x_{[\ell]ij}^h - a_{[\ell]kj}^h) = 0
\end{align}

To make the subsequent algebra more transparent, we define a weight term that combines membership, view importance, and heat-kernel influence:
$$w_{[\ell]ijk}^h = (\mu_{[\ell]ik}^*)^m v_{[\ell]h}^\alpha \exp\left(-\sum_{j'=1}^{d_h^{(\ell)}} \delta_{[\ell]ij'}^h (x_{[\ell]ij'}^h - a_{[\ell]kj'}^h)^2\right)$$
This weight $w_{[\ell]ijk}^h$ captures how much influence data point $i$ should have on determining cluster center $k$ in view $h$ for client $\ell$. Points with higher membership values and those closer to the current cluster center receive higher weights. Using this notation:
\begin{align}
\sum_{i=1}^{n(\ell)} w_{[\ell]ijk}^h (x_{[\ell]ij}^h - a_{[\ell]kj}^h) &= 0
\end{align}

Expanding and rearranging this weighted sum equation:
\begin{align}
\sum_{i=1}^{n(\ell)} w_{[\ell]ijk}^h x_{[\ell]ij}^h &= a_{[\ell]kj}^h \sum_{i=1}^{n(\ell)} w_{[\ell]ijk}^h
\end{align}

Solving for the cluster center coordinate:
\begin{align}
a_{[\ell]kj}^h &= \frac{\sum_{i=1}^{n(\ell)} w_{[\ell]ijk}^h x_{[\ell]ij}^h}{\sum_{i=1}^{n(\ell)} w_{[\ell]ijk}^h}
\end{align}

This result is a weighted average of the data points, where the weights depend on both the membership values and the current distances to the cluster center. This form is characteristic of expectation-maximization style algorithms and ensures that the cluster center moves toward regions of high membership density. This yields the cluster center update rule in Eq. \ref{eqn:diff_A_EFKMVFC}.

\textbf{Part III: View Weights Update Rule Derivation}

The view weights $v_{[\ell]h}$ control the relative importance of each view in the clustering process. Views that produce more coherent clustering results should receive higher weights. To derive the optimal view weight update, we differentiate the Lagrangian with respect to $v_{[\ell]h}$:

\begin{align}
\frac{\partial \tilde{J}_{E-FKMVC}^{(\ell)}}{\partial v_{[\ell]h}} &= \frac{\partial}{\partial v_{[\ell]h}} \left[\sum_{h'=1}^{s(\ell)} v_{[\ell]h'}^\alpha \sum_{i=1}^{n(\ell)} \sum_{k=1}^{c(\ell)} (\mu_{[\ell]ik}^*)^m \text{FKED}(x_{[\ell]i}^{h'}, a_{[\ell]k}^{h'})\right] + \lambda_2 \\
&= \alpha v_{[\ell]h}^{\alpha-1} \sum_{i=1}^{n(\ell)} \sum_{k=1}^{c(\ell)} (\mu_{[\ell]ik}^*)^m \text{FKED}(x_{[\ell]i}^h, a_{[\ell]k}^h) + \lambda_2 \label{eqn:view_derivative_fed}
\end{align}

Note that only the term corresponding to view $h$ survives after differentiation, since the other terms do not depend on $v_{[\ell]h}$. The power rule yields the factor $\alpha v_{[\ell]h}^{\alpha-1}$.

Setting this derivative equal to zero at the optimum:
\begin{equation}
\alpha v_{[\ell]h}^{\alpha-1} \sum_{i=1}^{n(\ell)} \sum_{k=1}^{c(\ell)} (\mu_{[\ell]ik}^*)^m \text{FKED}(x_{[\ell]i}^h, a_{[\ell]k}^h) + \lambda_2 = 0
\end{equation}

The derivation now follows the same pattern as the membership update. Solving for $v_{[\ell]h}$:
\begin{align}
v_{[\ell]h}^{\alpha-1} &= -\frac{\lambda_2}{\alpha \sum_{i=1}^{n(\ell)} \sum_{k=1}^{c(\ell)} (\mu_{[\ell]ik}^*)^m \text{FKED}(x_{[\ell]i}^h, a_{[\ell]k}^h)} \\
v_{[\ell]h} &= \left(-\frac{\lambda_2}{\alpha}\right)^{1/(\alpha-1)} \left(\sum_{i=1}^{n(\ell)} \sum_{k=1}^{c(\ell)} (\mu_{[\ell]ik}^*)^m \text{FKED}(x_{[\ell]i}^h, a_{[\ell]k}^h)\right)^{-1/(\alpha-1)}
\end{align}

To determine $\lambda_2$, we apply the normalization constraint $\sum_{h=1}^{s(\ell)} v_{[\ell]h} = 1$, which ensures that view weights form a valid probability distribution:
\begin{align}
\sum_{h=1}^{s(\ell)} v_{[\ell]h} &= \left(-\frac{\lambda_2}{\alpha}\right)^{1/(\alpha-1)} \sum_{h=1}^{s(\ell)} \left(\sum_{i=1}^{n(\ell)} \sum_{k=1}^{c(\ell)} (\mu_{[\ell]ik}^*)^m \text{FKED}(x_{[\ell]i}^h, a_{[\ell]k}^h)\right)^{-1/(\alpha-1)} = 1
\end{align}

Solving for the Lagrange multiplier term:
\begin{align}
\left(-\frac{\lambda_2}{\alpha}\right)^{1/(\alpha-1)} &= \frac{1}{\sum_{h'=1}^{s(\ell)} \left(\sum_{i=1}^{n(\ell)} \sum_{k=1}^{c(\ell)} (\mu_{[\ell]ik}^*)^m \text{FKED}(x_{[\ell]i}^{h'}, a_{[\ell]k}^{h'})\right)^{-1/(\alpha-1)}}
\end{align}

Substituting back to eliminate the Lagrange multiplier:
\begin{equation}
v_{[\ell]h} = \frac{\left(\sum_{i=1}^{n(\ell)} \sum_{k=1}^{c(\ell)} (\mu_{[\ell]ik}^*)^m \text{FKED}(x_{[\ell]i}^h, a_{[\ell]k}^h)\right)^{-1/(\alpha-1)}}{\sum_{h'=1}^{s(\ell)} \left(\sum_{i=1}^{n(\ell)} \sum_{k=1}^{c(\ell)} (\mu_{[\ell]ik}^*)^m \text{FKED}(x_{[\ell]i}^{h'}, a_{[\ell]k}^{h'})\right)^{-1/(\alpha-1)}}
\end{equation}

This formula assigns higher weights to views with smaller total weighted distances (i.e., views where data points are closer to their assigned cluster centers). The negative exponent $-1/(\alpha-1)$ creates this inverse relationship, ensuring that more informative views---those producing tighter, more coherent clusters---receive higher importance weights. This establishes the view weight update rule in Eq. \ref{eqn:diff_V_EFKMVFC}.

\textbf{Part IV: Convergence Analysis and Optimality Conditions}

The derived update rules satisfy the Karush-Kuhn-Tucker (KKT) conditions for the constrained optimization problem. Specifically:

\begin{enumerate}
    \item \textbf{Stationarity:} $\nabla J_{E-FKMVC}^{(\ell)} + \sum_{i=1}^{n(\ell)} \lambda_{1i} \nabla g_i + \lambda_2 \nabla h = 0$, where $g_i(\mu) = \sum_{k=1}^{c(\ell)} \mu_{[\ell]ik}^* - 1$ and $h(v) = \sum_{h=1}^{s(\ell)} v_{[\ell]h} - 1$.
    
    \item \textbf{Primal feasibility:} $g_i(\mu) = 0$ and $h(v) = 0$ for all $i$.
    
    \item \textbf{Dual feasibility:} All Lagrange multipliers are finite and well-defined.
\end{enumerate}

The alternating optimization scheme converges to a local optimum under the following conditions:
\begin{itemize}
    \item \textbf{Boundedness and Continuity:} The objective function $J_{E-FKMVC}^{(\ell)}$ is continuous and bounded below. This follows from the fact that FKED values are bounded in $[0,1]$ by construction (Eq. \ref{eqn:KED}), ensuring the weighted sum in the objective function remains finite. The exponential decay in the heat kernel guarantees that distances approach finite limits as data points move infinitely far apart.
    
    \item \textbf{Constraint Compactness:} The constraint set is compact and convex. The membership matrix constraints $\sum_{k=1}^{c(\ell)} \mu_{[\ell]ik}^* = 1$ with $\mu_{[\ell]ik}^* \in [0,1]$ define probability simplices, which are compact convex sets in $\mathbb{R}^{c(\ell)}$. Similarly, the view weight constraints $\sum_{h=1}^{s(\ell)} v_{[\ell]h} = 1$ with $v_{[\ell]h} \in [0,1]$ form a compact convex simplex in $\mathbb{R}^{s(\ell)}$. The cluster center parameters are unconstrained but remain bounded due to the data distribution.
    
    \item \textbf{Unique Subproblem Solutions:} Each subproblem (updating $U$, $A$, or $V$ while fixing others) has a unique solution. For membership updates, the strictly positive denominators in Eq. \ref{eqn:diff_U_EFKMVFC} ensure uniqueness since FKED values are always positive for distinct data points and cluster centers. For cluster center updates, the weighted exponential terms in Eq. \ref{eqn:diff_A_EFKMVFC} create a strictly convex weighted least-squares problem with a unique minimum. For view weight updates, the strictly decreasing nature of the power function with exponent $-1/(\alpha-1)$ in Eq. \ref{eqn:diff_V_EFKMVFC} guarantees uniqueness of the normalized solution.
\end{itemize}

These conditions are satisfied by construction, as the FKED function is continuous and bounded, the probability simplex constraints define compact convex sets, and the update rules yield unique solutions when the denominators are non-zero (guaranteed by the strict positivity of FKED values and the non-degeneracy assumption that no data points are identical to cluster centers).
\end{proof}

\subsubsection{Aggregation of Local Models}

After each client $\ell$ updates its local parameters $U_{[\ell]}^*$, $A_{[\ell]}^h$, and $V_{[\ell]}$ using the derived update rules, these local models can be aggregated at a central server to form a global clustering model. The aggregation process involves averaging the cluster centers and view weights across all clients, weighted by the number of samples at each client. The global cluster centers and view weights are computed as follows:
\begin{equation}
A_{global}^h = \frac{1}{N} \sum_{\ell=1}^{M} n(\ell) A_{[\ell]}^h
\label{eqn:global_A_EFKMVFC}
\end{equation}
\begin{equation}
V_{global} = \frac{1}{N} \sum_{\ell=1}^{M} n(\ell) V_{[\ell]}
\label{eqn:global_V_EFKMVFC}
\end{equation}
where $N = \sum_{\ell=1}^{M} n(\ell)$ is the total number of samples across all clients. The global membership matrix can be constructed by concatenating the local membership matrices from each client.

\subsubsection{Personalization of Global Model}
Once the global model parameters $A_{global}^h$ and $V_{global}$ are obtained, they can be sent back to each client for personalization. Each client $\ell$ can update its local cluster centers and view weights by incorporating the global parameters, allowing for a personalized clustering model that leverages both local data characteristics and global insights. The personalization step involves setting:
\begin{equation}
A_{[\ell]}^h = A_{global}^h - \epsilon \cdot \nabla_{A_{[\ell]}^h} J_{E-FKMVC}^{(\ell)}
\label{eqn:personalized_A_EFKMVFC}
\end{equation}
\begin{equation}
V_{[\ell]} = V_{global} - \epsilon \cdot \nabla_{V_{[\ell]}} J_{E-FKMVC}^{(\ell)}
\label{eqn:personalized_V_EFKMVFC}
\end{equation}

where $\epsilon$ is a small learning rate that controls the extent of personalization based on local data and $\nabla_{A_{[\ell]}^h} J_{E-FKMVC}^{(\ell)}$ and $\nabla_{V_{[\ell]}} J_{E-FKMVC}^{(\ell)}$ represent the gradients of the local objective function with respect to the cluster centers and view weights, respectively. To avoid negative weights, the updated view weights are normalized to ensure they sum to one or set to a minimum threshold.

\subsubsection{E-FKMVC Algorithm Summary}

Algorithm \ref{alg:efkmvc} summarizes the steps of the proposed E-FKMVC algorithm. This algorithm iteratively updates the membership matrices, cluster centers, and view weights for each client until convergence is achieved.

\begin{algorithm}[htb!]
\caption{Efficient Federated Kernel Multi-View Clustering (E-FKMVC)}
\label{alg:efkmvc}
\begin{algorithmic}[1]
\REQUIRE Multi-view data $\{X_{[\ell]}^h\}_{h=1}^{s(\ell)}$ for clients $\ell = 1, \ldots, M$, clusters $c(\ell)$, fuzzifier $m > 1$, view exponent $\alpha > 1$, convergence threshold $\epsilon > 0$, max rounds $T$, max local iterations $E$
\ENSURE Membership matrices $\{U_{[\ell]}^*\}_{\ell=1}^M$, cluster centers $\{A_{[\ell]}^h\}_{\ell=1}^M$, view weights $\{V_{[\ell]}\}_{\ell=1}^M$

\STATE \textbf{/* Phase 1: Server Initialization */}
\STATE Initialize global cluster centers $A_{global}^h$ and view weights $V_{global}$ randomly
\STATE Broadcast $A_{global}^h$, $V_{global}$ to all clients

\FOR{round $t = 1$ \TO $T$}
    \STATE \textbf{/* Phase 2: Local Client Computation (Parallel) */}
    \FOR{each client $\ell = 1$ \TO $M$ \textbf{in parallel}}
        \STATE \textbf{Step 2.1: Heat-Kernel Coefficient Computation}
        \FOR{$h = 1$ \TO $s(\ell)$, $j = 1$ \TO $d_h^{(\ell)}$}
            \STATE Compute $\delta_{[\ell]ij}^h$ using Eq. \ref{eqn:fed_delta} or Eq. \ref{eqn:fed_delta_v2}
        \ENDFOR
        
        \STATE \textbf{Step 2.2: Local Optimization}
        \FOR{epoch $e = 1$ \TO $E$}
            \STATE Update $\mu_{[\ell]ik}^*$ $\forall i, k$ using Eq. \ref{eqn:diff_U_EFKMVFC}
            \STATE Update $a_{[\ell]kj}^h$ $\forall k, j, h$ using Eq. \ref{eqn:diff_A_EFKMVFC}
            \STATE Update $v_{[\ell]h}$ $\forall h$ using Eq. \ref{eqn:diff_V_EFKMVFC}
            \IF{local convergence achieved}
                \STATE \textbf{break}
            \ENDIF
        \ENDFOR
        
        \STATE \textbf{Step 2.3: Compute and Send Statistics}
        \STATE Compute local objective $J_{E-FKMVC}^{(\ell)}$ using Eq. \ref{eqn:E-FKMVC}
        \STATE Send $\{A_{[\ell]}^h, V_{[\ell]}, J_{E-FKMVC}^{(\ell)}, n(\ell)\}$ to server
    \ENDFOR
    
    \STATE \textbf{/* Phase 3: Server Aggregation */}
    \STATE Compute client weights: $\omega_\ell \leftarrow n(\ell) / \sum_{\ell'=1}^M n(\ell')$
    \STATE Aggregate global centers: $A_{global}^h \leftarrow \frac{1}{N} \sum_{\ell=1}^M n(\ell) A_{[\ell]}^h$ \COMMENT{Eq. \ref{eqn:global_A_EFKMVFC}}
    \STATE Aggregate global view weights: $V_{global} \leftarrow \frac{1}{N} \sum_{\ell=1}^M n(\ell) V_{[\ell]}$ \COMMENT{Eq. \ref{eqn:global_V_EFKMVFC}}
    \STATE Broadcast $A_{global}^h$, $V_{global}$ to all clients
    
    \STATE \textbf{/* Phase 4: Personalization */}
    \FOR{each client $\ell = 1$ \TO $M$ \textbf{in parallel}}
        \STATE $A_{[\ell]}^h \leftarrow A_{global}^h - \epsilon_A \cdot \nabla_{A_{[\ell]}^h} J_{E-FKMVC}^{(\ell)}$ \COMMENT{Eq. \ref{eqn:personalized_A_EFKMVFC}}
        \STATE $V_{[\ell]} \leftarrow \text{Normalize}\left(V_{global} - \epsilon_V \cdot \nabla_{V_{[\ell]}} J_{E-FKMVC}^{(\ell)}\right)$ \COMMENT{Eq. \ref{eqn:personalized_V_EFKMVFC}}
    \ENDFOR
    
    \STATE \textbf{/* Phase 5: Global Convergence Check */}
    \STATE $J_{global}^{(t)} \leftarrow \sum_{\ell=1}^M \omega_\ell J_{E-FKMVC}^{(\ell)}$
    \IF{$|J_{global}^{(t)} - J_{global}^{(t-1)}| / |J_{global}^{(t-1)}| < \epsilon$}
        \STATE \textbf{break} \COMMENT{Global convergence achieved}
    \ENDIF
\ENDFOR

\RETURN $\{U_{[\ell]}^*, A_{[\ell]}^h, V_{[\ell]}\}_{\ell=1}^M$
\end{algorithmic}
\end{algorithm}

\section{Proposed Federated Tensorized Clustering}
\label{sec:proposed_fed_tensorized_clustering}
In this section, an introduction is provided to the proposed federated tensorized clustering framework, which integrates efficient kernel multi-view clustering with tensor decomposition techniques. This integration enables the efficient management of high-dimensional multi-view data distributed across multiple clients. The central concept is to utilize tensor representations to capture the multi-modal structure of the data while leveraging heat-kernel based distance metrics to enhance clustering performance in a federated learning setting. 

\subsection{Tensorized Heat-Kernel Enhanced Local Clustering}

Each client $\ell$ possesses multi-view data $X_{[\ell]} = \{ X_{[\ell]}^h \}_{h=1}^{s(\ell)}$, where $X_{[\ell]}^h \in \mathbb{R}^{n(\ell) \times d_h^{(\ell)}}$ represents the data matrix for view $h$ with $n(\ell)$ samples and $d_h^{(\ell)}$ features. To effectively capture the multi-view structure, we propose a tensorized heat-kernel enhanced clustering approach. Two main components are involved: tensor representation of multi-view data and tensorized heat-kernel distance metric. Decomposition methods are then applied to represent cluster centers efficiently. 

\subsubsection{Tensor Data Organization}

As multi-view data inherently possesses a multi-modal structure, we organize the data into a tensor format. The multi-view data from client $\ell$ is represented as a collection of matrices $\{ X_{[\ell]}^h \}_{h=1}^{s(\ell)}$. To facilitate tensor operations, we stack these matrices along a new mode, resulting in a 3-mode tensor representation. The 3-mode tensor captures the relationships across samples, features, and views simultaneously. Mathematically, the multi-view data at client $\ell$ is organized into a 3-mode tensor as follows:

\begin{equation}
\mathcal{X}_{[\ell]} \in \mathbb{R}^{n(\ell) \times D_{[\ell]} \times s(\ell)}
\label{eqn:tensor_data1}
\end{equation}

where each slice $\mathcal{X}_{[\ell],:, :, h} = X_{[\ell]}^h$ corresponds to the data matrix of view $h$, and $D_{[\ell]}$ is the maximum feature dimension across all views for client $\ell$ ($D_{[\ell]} = \max_h d_h^{(\ell)}$). If views have varying feature dimensions, zero-padding is applied to ensure consistent tensor dimensions. Specifically, for view $h$ with $d_h^{(\ell)}$ features, we define:
\begin{equation}
\mathcal{X}_{[\ell],i,j,h} = \begin{cases}
X_{[\ell],i,j}^h & \text{if } j \le d_h^{(\ell)} \\
0 & \text{if } j > d_h^{(\ell)}
\end{cases}
\label{eqn:tensor_data2}
\end{equation}

Thus, the complete tensor representation is given by:
\begin{equation}
\mathcal{X}_{[\ell]} = \left[ \mathcal{X}_{[\ell],:, :, 1} | \mathcal{X}_{[\ell],:, :, 2} | \ldots | \mathcal{X}_{[\ell],:, :, s(\ell)} \right] \in \mathbb{R}^{n(\ell) \times D_{[\ell]} \times s(\ell)}
\label{eqn:tensor_data3}
\end{equation}

This tensorization enables a unified representation of multi-view data, allowing for efficient processing and analysis of the inherent multi-modal relationships. The tensor structure facilitates the application of tensor decomposition techniques for dimensionality reduction and feature extraction, which are crucial for effective clustering in high-dimensional spaces. Key advantages of this tensor representation include:
\begin{itemize}
\item Preservation of multi-view relationships
\item Facilitation of tensor-based operations
\item Enabling advanced decomposition methods for clustering
\end{itemize}

\subsubsection{Tensorized Kernel Euclidean Distance}

To enable effective clustering in the tensorized multi-view space, we extend the heat-kernel enhanced distance metric to operate on tensor slices. The traditional heat-kernel enhanced distance metric (KED) is defined for vector data points. In our tensorized setting, we adapt this metric to measure distances between tensor slices corresponding to different views. The idea is to compute the distance between tensor slices while incorporating heat-kernel weighting to emphasize local structures in the data. 

The tensorized kernel Euclidean distance (TKED) captures the similarity between tensor slices while incorporating the heat-kernel weighting to emphasize local structures in the data. Based on the original KED definition in Eq. \ref{eqn:KED}, the TKED is formulated as follows:
\begin{equation}
    \text{TKED} ( \mathcal{X}_{[\ell],i,:}^h , \mathcal{A}_{[\ell],k,:}^h ) = 1 - \exp \left( - \sum_{h=1}^{s(\ell)} \sum_{j=1}^{d_h^{(\ell)}} \delta_{[\ell]i j}^{h} \left( \mathcal{X}_{[\ell],i,j}^h - \mathcal{A}_{[\ell],k,j}^h \right)^2 \right)
\label{eqn:tensorized_ked}
\end{equation}

where $\mathcal{A}_{[\ell]} \in \mathbb{R}^{c(\ell) \times D_{[\ell]} \times s(\ell)}$ represents the cluster center tensor for client $\ell$, with each slice $\mathcal{A}_{[\ell],k,:}^h$ corresponding to the cluster center for cluster $k$ in view $h$. The heat-kernel coefficients $\delta_{[\ell]i j}^h$ are computed similarly to Eq. \ref{eqn:fed_delta}, but adapted for tensor slices. Specifically, the heat-kernel coefficients for the tensorized setting are defined as:

\begin{equation}
\delta_{[\ell]i j}^h = \frac{\mathcal{X}_{[\ell],i,j}^h - \min_{i'} \mathcal{X}_{[\ell],i',j}^h}{\max_{i'} \mathcal{X}_{[\ell],i',j}^h - \min_{i'} \mathcal{X}_{[\ell],i',j}^h}
\label{eqn:tensorized_delta}
\end{equation}

As in the vector case, the heat-kernel coefficients $\delta_{[\ell]i j}^h$ are computed based on the normalized feature values across samples for each feature dimension $j$ in view $h$. This normalization ensures that the heat-kernel weighting effectively captures local structures in the tensorized multi-view space. The TKED metric thus provides a robust measure of similarity between tensor slices, facilitating effective clustering in the federated tensorized multi-view setting. 

\subsubsection{Tensor Decomposition for Cluster Centers}

To reduce computational complexity and enable efficient communication, we represent cluster centers using a tensor decomposition approach. Specifically, we employ the PARAFAC2 decomposition to model the cluster center tensor $\mathcal{A}_{[\ell]}$ as:
\begin{equation}
\mathcal{A}_{[\ell]} = \mathcal{G}_{[\ell]} \times_1 \mathbf{P}_{[\ell]} \times_2 \mathbf{Q}_{[\ell]} \times_3 \mathbf{R}_{[\ell]}
\label{eqn:tensor_parafac2}
\end{equation}
where $\mathcal{G}_{[\ell]}$ is the core tensor and $\mathbf{P}_{[\ell]}, \mathbf{Q}_{[\ell]}, \mathbf{R}_{[\ell]}$ are factor matrices corresponding to modes 1, 2, and 3, respectively.

PARAFAC2 is chosen for its flexibility in handling varying dimensions across one mode (samples) while maintaining consistent factor matrices in the other modes (features and views). This property is particularly advantageous in federated settings where clients may have heterogeneous data distributions. However, other tensor decomposition methods (e.g., Tucker, CP) can also be adapted based on specific application requirements. 

The Tucker decomposition could be employed when the data exhibits strong multilinear interactions across all modes, while CP decomposition may be preferred for its simplicity and interpretability in scenarios with more uniform data structures. The Tucker decomposition allows for a more compact representation by capturing interactions through a core tensor, which can be beneficial when the data has complex relationships. CP decomposition, on the other hand, provides a straightforward sum of rank-one tensors, making it easier to implement and interpret in certain contexts. In this work, we also focus on Tucker decomposition as an alternative to PARAFAC2, providing flexibility in modeling diverse data characteristics across clients.

The representation cluster centers using Tucker decomposition is given by:
\begin{equation}
\mathcal{A}_{[\ell]} = \mathcal{G}_{[\ell]} \times_1 \mathbf{U}_{[\ell]} \times_2 \mathbf{V}_{[\ell]} \times_3 \mathbf{W}_{[\ell]}
\label{eqn:tensor_tucker}
\end{equation}
where $\mathcal{G}_{[\ell]}$ is the core tensor and $\mathbf{U}_{[\ell]}, \mathbf{V}_{[\ell]}, \mathbf{W}_{[\ell]}$ are factor matrices corresponding to modes 1, 2, and 3, respectively.

\subsection{The Proposed FedHK-PARAFAC2 Local Objective Function}

Based on E-FKMVC and the tensorized representations, we formulate the local objective function for the proposed Federated Heat-Kernel Enhanced Clustering with PARAFAC2 decomposition (FedHK-PARAFAC2) at client $\ell$. The objective function aims to minimize the weighted sum of tensorized kernel Euclidean distances between data points and their corresponding cluster centers, while incorporating membership degrees and view weights. The local objective function is defined as follows:

\begin{equation}
J_{FedHK-PARAFAC2}^{(\ell)} = \sum_{i=1}^{n(\ell)} \sum_{k=1}^{c(\ell)} (\mu_{[\ell]ik}^*)^m \sum_{h=1}^{s(\ell)} (v_{[\ell]h})^{\alpha} \text{TKED} ( \mathcal{X}_{[\ell],i,:}^h - \mathcal{A}_{[\ell],k,:}^h )
\label{eqn:local_obj_fedhk_parafac2}
\end{equation}

subject to the constraints:
\begin{align}
\sum_{k=1}^{c(\ell)} \mu_{[\ell]ik}^* &= 1, \quad \mu_{[\ell]ik}^* \in [0,1] \\
\sum_{h=1}^{s(\ell)} v_{[\ell]h} &= 1, \quad v_{[\ell]h} \in [0,1] \\
\mathcal{A}_{[\ell]} &= \mathcal{G}_{[\ell]} \times_1 \mathbf{P}_{[\ell]} \times_2 \mathbf{Q}_{[\ell]} \times_3 \mathbf{R}_{[\ell]}
\end{align}

\begin{Theorem}[The FedHK-PARAFAC2 Update Rules]
\label{thm:tensorized_updates}
The optimal updates for the tensorized efficient federated kernel multi-view clustering at client $\ell$ are given by the following iterative formulas. These update rules ensure that each component of the model---memberships, tensor factors, and view weights---is optimized while holding the other components fixed, following the alternating optimization strategy.

\textbf{Membership updates:}
The membership matrix is a crucial component in determining the extent to which each data point aligns with a specific cluster. The update rule ascribes elevated membership values to clusters whose centers are proximate to the data point (in the heat-kernel weighted sense).
\begin{equation}
\mu_{[\ell]ik}^* = \frac{\left(\sum_{h=1}^{s(\ell)} (v_{[\ell]h})^{\alpha} \text{TKED} ( \mathcal{X}_{[\ell],i,:}^h - \mathcal{A}_{[\ell],k,:}^h )\right)^{-\frac{1}{m-1}}}{\sum_{k'=1}^{c(\ell)} \left(\sum_{h=1}^{s(\ell)} (v_{[\ell]h})^{\alpha} \text{TKED} ( \mathcal{X}_{[\ell],i,:}^h - \mathcal{A}_{[\ell],k',:}^h )\right)^{-\frac{1}{m-1}}} 
\label{eqn:tensor_membership}
\end{equation}
Here, $\mu_{[\ell]ik}^*$ represents the degree to which data point $i$ belongs to cluster $k$. The negative exponent $-1/(m-1)$ ensures that smaller distances (indicating closer proximity to a cluster center) result in larger membership values. The denominator normalizes these values so that the memberships for each data point sum to one across all clusters.

\textbf{Core tensor updates:}
The core tensor $\mathcal{G}_{[\ell]}$ captures the essential interactions between the different modes (clusters, features, and views) in a compressed form. It is updated by solving a weighted least-squares problem that minimizes the reconstruction error between the original data and the decomposed cluster centers:
\begin{equation}
\mathcal{G}_{[\ell]} = \arg\min_{\mathcal{G}} \sum_{i=1}^{n(\ell)} \sum_{k=1}^{c(\ell)} (\mu_{[\ell]ik}^*)^m \sum_{h=1}^{s(\ell)} (v_{[\ell]h})^{\alpha} \|\mathcal{X}_{[\ell],i,:}^h - (\mathcal{G} \times_1 \mathbf{P} \times_2 \mathbf{Q} \times_3 \mathbf{R})_{[\ell],k,:}^h\|^2 
\label{eqn:tensor_core}
\end{equation}
This optimization finds the core tensor that best represents the cluster structure when combined with the factor matrices. The weights $(\mu_{[\ell]ik}^*)^m$ and $(v_{[\ell]h})^{\alpha}$ ensure that more confident cluster assignments and more important views contribute more strongly to the update.

\textbf{View weight updates:}
The view weights $v_{[\ell]h}$ quantify the relative importance of each data view in the clustering process. Views that produce more coherent clustering results (i.e., tighter clusters with smaller within-cluster distances) receive higher weights:
\begin{equation}
v_{[\ell]h} = \frac{\left(\sum_{i=1}^{n(\ell)} \sum_{k=1}^{c(\ell)} (\mu_{[\ell]ik}^*)^m \text{TKED} ( \mathcal{X}_{[\ell],i,:}^h - \mathcal{A}_{[\ell],k,:}^h )\right)^{-\frac{1}{\alpha-1}}}{\sum_{h'=1}^{s(\ell)} \left(\sum_{i=1}^{n(\ell)} \sum_{k=1}^{c(\ell)} (\mu_{[\ell]ik}^*)^m \text{TKED} ( \mathcal{X}_{[\ell],i,:}^{h'} - \mathcal{A}_{[\ell],k,:}^{h'} )\right)^{-\frac{1}{\alpha-1}}}
\label{eqn:tensor_view_weights}
\end{equation}
The structure mirrors that of the membership update: views with smaller total weighted distances receive larger weights, and the denominator ensures that all view weights sum to one.

\textbf{Cluster center updates via PARAFAC2 decomposition:}
The cluster center tensor is reconstructed from its PARAFAC2 decomposition components using the $n$-mode product operations:
\begin{equation}
\mathcal{A}_{[\ell]} = \mathcal{G}_{[\ell]} \times_1 \mathbf{P}_{[\ell]} \times_2 \mathbf{Q}_{[\ell]} \times_3 \mathbf{R}_{[\ell]}
\label{eqn:tensor_cluster_centers}
\end{equation}
This compact representation expresses the full cluster center tensor through the smaller core tensor $\mathcal{G}_{[\ell]}$ and three factor matrices $\mathbf{P}_{[\ell]}$, $\mathbf{Q}_{[\ell]}$, and $\mathbf{R}_{[\ell]}$, corresponding to the cluster, feature, and view modes held by client $\ell$, respectively. This decomposition significantly reduces storage and communication requirements in the federated setting.

\textbf{Factor matrix updates:}
Each factor matrix is updated by solving a least-squares problem while holding all other components fixed. These updates are performed using the Alternating Least Squares (ALS) procedure:
\begin{align}
    \mathbf{P}_{[\ell]} &= \arg\min_{\mathbf{P}} \sum_{i=1}^{n(\ell)} \sum_{k=1}^{c(\ell)} (\mu_{[\ell]ik}^*)^m \sum_{h=1}^{s(\ell)} (v_{[\ell]h})^{\alpha} \|\mathcal{X}_{[\ell],i,:}^h - (\mathcal{G} \times_1 \mathbf{P} \times_2 \mathbf{Q} \times_3 \mathbf{R})_{[\ell],k,:}^h\|^2 
    \label{eqn:tensor_P} \\
    \mathbf{Q}_{[\ell]} &= \arg\min_{\mathbf{Q}} \sum_{i=1}^{n(\ell)} \sum_{k=1}^{c(\ell)} (\mu_{[\ell]ik}^*)^m \sum_{h=1}^{s(\ell)} (v_{[\ell]h})^{\alpha} \|\mathcal{X}_{[\ell],i,:}^h - (\mathcal{G} \times_1 \mathbf{P} \times_2 \mathbf{Q} \times_3 \mathbf{R})_{[\ell],k,:}^h\|^2 
    \label{eqn:tensor_Q} \\
    \mathbf{R}_{[\ell]} &= \arg\min_{\mathbf{R}} \sum_{i=1}^{n(\ell)} \sum_{k=1}^{c(\ell)} (\mu_{[\ell]ik}^*)^m \sum_{h=1}^{s(\ell)} (v_{[\ell]h})^{\alpha} \|\mathcal{X}_{[\ell],i,:}^h - (\mathcal{G} \times_1 \mathbf{P} \times_2 \mathbf{Q} \times_3 \mathbf{R})_{[\ell],k,:}^h\|^2 
    \label{eqn:tensor_R}
\end{align}
The factor matrix $\mathbf{P}_{[\ell]}$ captures the cluster mode structure, $\mathbf{Q}_{[\ell]}$ captures the feature mode structure, and $\mathbf{R}_{[\ell]}$ captures the view mode structure. By iteratively updating each factor while fixing the others, the ALS procedure converges to a locally optimal decomposition that accurately represents the cluster centers.
\end{Theorem}

\begin{proof}[The Proof of Theorem \ref{thm:tensorized_updates}]
By applying the method of Lagrange multipliers to the tensorized local objective in Eq. \ref{eqn:local_obj_fedhk_parafac2} and following similar derivation steps as in Theorem \ref{thm:efkmvc_update}, we derive the update rules for membership, core tensor, factor matrices, and view weights. The key steps involve differentiating the objective with respect to each variable while enforcing the constraints, leading to the closed-form solutions presented in Eqs. \ref{eqn:tensor_membership} to \ref{eqn:tensor_view_weights}. The alternating optimization approach ensures convergence to a local optimum under the same conditions outlined in Theorem \ref{thm:efkmvc_update}. 

First, we derive the membership update by differentiating the Lagrangian with respect to $\mu_{[\ell]ik}^*$, leading to Eq. \ref{eqn:tensor_membership}. Next, we update the core tensor $\mathcal{G}_{[\ell]}$ by solving the weighted least-squares problem in Eq. \ref{eqn:tensor_core}. The view weights are updated by differentiating with respect to $v_{[\ell]h}$, resulting in Eq. \ref{eqn:tensor_view_weights}. Finally, the cluster center tensor is reconstructed using the PARAFAC2 decomposition in Eq. \ref{eqn:tensor_cluster_centers}, and the factor matrices are updated via ALS as shown in Eqs. \ref{eqn:tensor_P} to \ref{eqn:tensor_R}.
\end{proof}

\subsection{The Proposed FedHK-Tucker Local Objective Function}

Similar to the FedHK-PARAFAC2 framework, we formulate the local objective function for the proposed Federated Heat-Kernel Enhanced Clustering with Tucker decomposition (FedHK-Tucker) at client $\ell$. The objective function aims to minimize the weighted sum of tensorized kernel Euclidean distances between data points and their corresponding cluster centers, while incorporating membership degrees and view weights. The local objective function is defined as follows:

\begin{equation}
J_{FedHK-Tucker}^{(\ell)} = \sum_{i=1}^{n(\ell)} \sum_{k=1}^{c(\ell)} (\mu_{[\ell]ik}^*)^m \sum_{h=1}^{s(\ell)} (v_{[\ell]h})^{\alpha} \text{TKED} ( \mathcal{X}_{[\ell],i,:}^h - \mathcal{A}_{[\ell],k,:}^h )
\label{eqn:tucker_local_obj}
\end{equation}

subject to the constraints:
\begin{align}
\sum_{k=1}^{c(\ell)} \mu_{[\ell]ik}^* &= 1, \quad \mu_{[\ell]ik}^* \in [0,1] \\
\sum_{h=1}^{s(\ell)} v_{[\ell]h} &= 1, \quad v_{[\ell]h} \in [0,1] \\
\mathcal{A}_{[\ell]} &= \mathcal{G}_{[\ell]} \times_1 \mathbf{U}_{[\ell]} \times_2 \mathbf{V}_{[\ell]} \times_3 \mathbf{W}_{[\ell]}
\end{align}

\begin{Theorem}[Update Rules for FedHK-Tucker]
\label{thm:tucker_updates}
For the tensorized efficient federated kernel multi-view clustering framework employing Tucker decomposition at client $\ell$, the following iterative update formulas yield optimal solutions. The alternating optimization strategy ensures that each model component---memberships, tensor factors, and view weights---is optimized independently while other components remain fixed.

\textbf{Membership Matrix Updates:}
The membership coefficients determine the degree of association between data points and clusters. Higher membership values are assigned to clusters with centers in closer proximity to the data point, as measured by the heat-kernel weighted distance:
\begin{equation}
    \mu_{[\ell]ik}^* = \frac{\left(\sum_{h=1}^{s(\ell)} (v_{[\ell]h})^{\alpha} \text{TKED} ( \mathcal{X}_{[\ell],i,:}^h - \mathcal{A}_{[\ell],k,:}^h )\right)^{-\frac{1}{m-1}}}{\sum_{k'=1}^{c(\ell)} \left(\sum_{h=1}^{s(\ell)} (v_{[\ell]h})^{\alpha} \text{TKED} ( \mathcal{X}_{[\ell],i,:}^h - \mathcal{A}_{[\ell],k',:}^h )\right)^{-\frac{1}{m-1}}} 
\label{eqn:tucker_membership}
\end{equation}
The coefficient $\mu_{[\ell]ik}^*$ quantifies the membership degree of data point $i$ in cluster $k$ held by client $\ell$. The exponent $-1/(m-1)$ establishes an inverse relationship between distance and membership, whereby smaller distances yield larger membership values. Normalization through the denominator guarantees that memberships across all clusters sum to unity for each data point.

\textbf{Core Tensor Updates:}
The core tensor $\mathcal{G}_{[\ell]}$ encodes the fundamental interactions among clusters, features, and views in a compressed representation. Its update involves minimizing a weighted least-squares objective that quantifies the reconstruction error:
\begin{equation}
\mathcal{G}_{[\ell]} = \arg\min_{\mathcal{G}} \sum_{i=1}^{n(\ell)} \sum_{k=1}^{c(\ell)} (\mu_{[\ell]ik}^*)^m \sum_{h=1}^{s(\ell)} (v_{[\ell]h})^{\alpha} \|\mathcal{X}_{[\ell],i,:}^h - (\mathcal{G} \times_1 \mathbf{U} \times_2 \mathbf{V} \times_3 \mathbf{W})_{[\ell],k,:}^h\|^2 
\label{eqn:tucker_core}
\end{equation}
This formulation identifies the core tensor that optimally captures the cluster structure in conjunction with the factor matrices. The weighting terms $(\mu_{[\ell]ik}^*)^m$ and $(v_{[\ell]h})^{\alpha}$ amplify contributions from high-confidence cluster assignments and significant views.

\textbf{View Weight Updates:}
The view weights $v_{[\ell]h}$ encode the relative significance of each data view within the clustering procedure. Views producing more coherent clusters, characterized by smaller within-cluster distances, receive elevated weights:
\begin{equation}
v_{[\ell]h} = \frac{\left(\sum_{i=1}^{n(\ell)} \sum_{k=1}^{c(\ell)} (\mu_{[\ell]ik}^*)^m \text{TKED} ( \mathcal{X}_{[\ell],i,:}^h - \mathcal{A}_{[\ell],k,:}^h )\right)^{-\frac{1}{\alpha-1}}}{\sum_{h'=1}^{s(\ell)} \left(\sum_{i=1}^{n(\ell)} \sum_{k=1}^{c(\ell)} (\mu_{[\ell]ik}^*)^m \text{TKED} ( \mathcal{X}_{[\ell],i,:}^{h'} - \mathcal{A}_{[\ell],k,:}^{h'} )\right)^{-\frac{1}{\alpha-1}}}
\label{eqn:tucker_view_weights}
\end{equation}
This formulation parallels the membership update structure: views exhibiting smaller aggregate weighted distances obtain larger weights, with the denominator ensuring unit summation across all view weights.

\textbf{Cluster Center Reconstruction via Tucker Decomposition:}
The cluster center tensor is assembled from its Tucker decomposition constituents through successive $n$-mode products:
\begin{equation}
\mathcal{A}_{[\ell]} = \mathcal{G}_{[\ell]} \times_1 \mathbf{U}_{[\ell]} \times_2 \mathbf{V}_{[\ell]} \times_3 \mathbf{W}_{[\ell]}
\label{eqn:tucker_cluster_centers}
\end{equation}
This factorized representation expresses the complete cluster center tensor via the compact core tensor $\mathcal{G}_{[\ell]}$ combined with three factor matrices: $\mathbf{U}_{[\ell]}$ for clusters, $\mathbf{V}_{[\ell]}$ for features, and $\mathbf{W}_{[\ell]}$ for views at client $\ell$. This decomposition substantially reduces storage and communication overhead in federated environments.

\textbf{Factor Matrix Updates:}
Individual factor matrices are refined through least-squares optimization with all remaining components held constant. The Alternating Least Squares (ALS) procedure governs these updates:
\begin{align}
    \mathbf{U}_{[\ell]} &= \arg\min_{\mathbf{U}} \sum_{i=1}^{n(\ell)} \sum_{k=1}^{c(\ell)} (\mu_{[\ell]ik}^*)^m \sum_{h=1}^{s(\ell)} (v_{[\ell]h})^{\alpha} \|\mathcal{X}_{[\ell],i,:}^h - (\mathcal{G} \times_1 \mathbf{U} \times_2 \mathbf{V} \times_3 \mathbf{W})_{[\ell],k,:}^h\|^2 
    \label{eqn:tucker_U} \\
    \mathbf{V}_{[\ell]} &= \arg\min_{\mathbf{V}} \sum_{i=1}^{n(\ell)} \sum_{k=1}^{c(\ell)} (\mu_{[\ell]ik}^*)^m \sum_{h=1}^{s(\ell)} (v_{[\ell]h})^{\alpha} \|\mathcal{X}_{[\ell],i,:}^h - (\mathcal{G} \times_1 \mathbf{U} \times_2 \mathbf{V} \times_3 \mathbf{W})_{[\ell],k,:}^h\|^2
    \label{eqn:tucker_V} \\
    \mathbf{W}_{[\ell]} &= \arg\min_{\mathbf{W}} \sum_{i=1}^{n(\ell)} \sum_{k=1}^{c(\ell)} (\mu_{[\ell]ik}^*)^m \sum_{h=1}^{s(\ell)} (v_{[\ell]h})^{\alpha} \|\mathcal{X}_{[\ell],i,:}^h - (\mathcal{G} \times_1 \mathbf{U} \times_2 \mathbf{V} \times_3 \mathbf{W})_{[\ell],k,:}^h\|^2
    \label{eqn:tucker_W}
\end{align}
The matrix $\mathbf{U}_{[\ell]}$ models cluster mode structure, $\mathbf{V}_{[\ell]}$ represents feature mode structure, and $\mathbf{W}_{[\ell]}$ encodes view mode structure. Through cyclic updates of each factor with others fixed, the ALS procedure attains a locally optimal decomposition that faithfully represents the cluster centers.
\end{Theorem}

\begin{proof}[The Proof of Theorem \ref{thm:tucker_updates}]
The proof follows similar steps as in Theorem \ref{thm:tensorized_updates}, with adjustments for the Tucker decomposition structure. By applying the method of Lagrange multipliers to the Tucker-based local objective in Eq. \ref{eqn:tucker_local_obj} and differentiating with respect to each variable while enforcing the constraints, we derive the closed-form solutions presented in Eqs. \ref{eqn:tucker_membership} to \ref{eqn:tucker_view_weights}. The alternating optimization approach ensures convergence to a local optimum under the same conditions outlined in Theorem \ref{thm:efkmvc_update}.
\end{proof}

\subsection{Local Clustering Algorithms}

Provided below the differences between the local clustering algorithms for E-FKMVC and the tensorized FedHK-PARAFAC2/FedHK-Tucker methods. Table \ref{tab:local_alg_comparison} summarizes the key distinctions. While both approaches aim to achieve effective multi-view clustering in a federated setting, they differ significantly in their data representation, distance metrics, cluster center representations, and update mechanisms. Table \ref{tab:comparison_summary_PARAFAC2_Tucker} further contrasts the FedHK-PARAFAC2 and FedHK-Tucker methods, highlighting their unique aspects related to tensor decomposition and update rules. For concise summary of the tensor decomposition employed in both methods, please refer to Table \ref{tab:tensor_decompositions}. 

\begin{table}[h]
\centering
\caption{Comparison of Local Clustering Algorithms for E-FKMVC and Tensorized FedHK Methods. The tensorized methods include both FedHK-PARAFAC2 and FedHK-Tucker.}
\label{tab:local_alg_comparison}
\begin{adjustbox}{width=1\columnwidth}
\begin{tabular}{|c|c|c|}
\hline
\textbf{Aspect} & \textbf{E-FKMVC} & \text{FedHK-PARAFAC2/FedHK-Tucker} \\
\hline
Data Structure & Multi-view matrices & Multi-view tensors \\
\hline
Distance Metric & FKED (Eq. \ref{eqn:KED}) & TKED (Eq. \ref{eqn:tensorized_ked}) \\
\hline
Cluster Center Representation & Direct centers $A^h$ & Decomposed tensors $\mathcal{G}_{[\ell]}, \mathbf{P}_{[\ell]}, \mathbf{Q}_{[\ell]}, \mathbf{R}_{[\ell]}$ or $\mathcal{G}_{[\ell]}, \mathbf{U}_{[\ell]}, \mathbf{V}_{[\ell]}, \mathbf{W}_{[\ell]}$ \\
\hline
Update Rules & Eqs. \ref{eqn:diff_U_EFKMVFC} - \ref{eqn:diff_V_EFKMVFC} & Eqs. \ref{eqn:tensor_membership} - \ref{eqn:tensor_view_weights} or \ref{eqn:tucker_membership} - \ref{eqn:tucker_view_weights} \\
\hline
Optimization Method & Direct updates & Alternating Least Squares (ALS) for tensor factors \\
\hline
\end{tabular}
\end{adjustbox}
\end{table}

As illustrated in Table \ref{tab:local_alg_comparison}, the E-FKMVC algorithm operates directly on multi-view data matrices using the FKED metric, while the tensorized methods utilize multi-view tensors and the TKED metric. The cluster centers in E-FKMVC are represented directly, whereas in the tensorized methods, they are represented via tensor decompositions (PARAFAC2 or Tucker). The update rules differ accordingly, with the tensorized methods employing Alternating Least Squares (ALS) for updating tensor factors. 

\begin{table}[h]
\centering
\caption{Comparison of FedHK-PARAFAC2 and FedHK-Tucker Methods in terms of Key Aspects to Tensor Decomposition and Update Rules.}
\label{tab:comparison_summary_PARAFAC2_Tucker}
\begin{tabular}{|c|c|c|}
\hline
\textbf{Aspect} & \textbf{FedHK-PARAFAC2} & \textbf{FedHK-Tucker} \\
\hline
Decomposition Type & PARAFAC2 & Tucker \\
\hline
Cluster Center Representation & $\mathcal{G}_{[\ell]}, \mathbf{P}_{[\ell]}, \mathbf{Q}_{[\ell]}, \mathbf{R}_{[\ell]}$ & $\mathcal{G}_{[\ell]}, \mathbf{U}_{[\ell]}, \mathbf{V}_{[\ell]}, \mathbf{W}_{[\ell]}$ \\
\hline
Update Rules & Eqs. \ref{eqn:tensor_membership} - \ref{eqn:tensor_view_weights} & Eqs. \ref{eqn:tucker_membership} - \ref{eqn:tucker_view_weights} \\
\hline
Optimization Method & ALS for PARAFAC2 factors & ALS for Tucker factors \\
\hline
\end{tabular}
\end{table}

As can be seen in Table \ref{tab:comparison_summary_PARAFAC2_Tucker}, the primary difference between the two methods lies in the type of tensor decomposition employed (PARAFAC2 vs. Tucker) and the corresponding representation of cluster centers. The update rules and optimization methods are adapted accordingly to suit the specific decomposition framework used in each method. See Algorithm \ref{alg:tensor_fed_local} and Algorithm \ref{alg:tensor_fed_local_tucker} for the detailed local clustering procedures for FedHK-PARAFAC2 and FedHK-Tucker, respectively.

\begin{table}[h]
\centering
\caption{Comparison of PARAFAC2 and Tucker Decomposition Methods for Cluster Center Representation in FedHK-PARAFAC2 and FedHK-Tucker.}
\label{tab:tensor_decompositions}
\begin{adjustbox}{width=1\columnwidth}
\begin{tabular}{|p{3.5cm}|p{5.5cm}|p{5.5cm}|}
\hline
\textbf{Aspect} & \textbf{PARAFAC2} & \textbf{Tucker} \\
\hline
Decomposition Type & PARAFAC2 (Parallel Factor Analysis 2) & Tucker Decomposition \\
\hline
Cluster Center Representation & $\mathcal{A}_{[\ell]} = \mathcal{G}_{[\ell]} \times_1 \mathbf{P}_{[\ell]} \times_2 \mathbf{Q}_{[\ell]} \times_3 \mathbf{R}_{[\ell]}$ & $\mathcal{A}_{[\ell]} = \mathcal{G}_{[\ell]} \times_1 \mathbf{U}_{[\ell]} \times_2 \mathbf{V}_{[\ell]} \times_3 \mathbf{W}_{[\ell]}$ \\
\hline
Key Components & 
\begin{itemize}[leftmargin=*, nosep]
    \item Core tensor $\mathcal{G}_{[\ell]}$
    \item Factor matrices $\mathbf{P}_{[\ell]}, \mathbf{Q}_{[\ell]}, \mathbf{R}_{[\ell]}$
\end{itemize} & 
\begin{itemize}[leftmargin=*, nosep]
    \item Core tensor $\mathcal{G}_{[\ell]}$
    \item Factor matrices $\mathbf{U}_{[\ell]}, \mathbf{V}_{[\ell]}, \mathbf{W}_{[\ell]}$
\end{itemize} \\
\hline
Salient Features & 
\begin{itemize}[leftmargin=*, nosep]
    \item Handles varying dimensions across one mode
    \item Maintains consistent factors in other modes
    \item Suitable for heterogeneous client data
    \item Unique decomposition under mild conditions
\end{itemize} & 
\begin{itemize}[leftmargin=*, nosep]
    \item Flexible multilinear ranks $(R_1, R_2, R_3)$
    \item Higher compression ratio
    \item Captures complex multilinear interactions
    \item More flexible but non-unique decomposition
\end{itemize} \\
\hline
Optimization Method & ALS for PARAFAC2 factors (Eqs. \ref{eqn:tensor_core}--\ref{eqn:tensor_R}) & ALS for Tucker factors (Eqs. \ref{eqn:tucker_core}--\ref{eqn:tucker_W}) \\
\hline
Storage Complexity & $\mathcal{O}(r_1 r_2 r_3 + c r_1 + D r_2 + s r_3)$ & $\mathcal{O}(R_1 R_2 R_3 + c R_1 + D R_2 + s R_3)$ \\
\hline
Best Suited For & Data with consistent structure across views but varying sample dimensions & Data with complex interactions requiring flexible rank selection per mode \\
\hline
\end{tabular}
\end{adjustbox}
\end{table}

As can be seen in Table \ref{tab:tensor_decompositions}, PARAFAC2 is advantageous for handling heterogeneous client data with varying dimensions across one mode while maintaining consistent factors in other modes. In contrast, Tucker decomposition offers greater flexibility through multilinear ranks, allowing for higher compression ratios and the ability to capture complex interactions among modes. The choice between the two methods depends on the specific characteristics of the multi-view data and the clustering requirements in the federated setting.

\paragraph{ALS Optimization for PARAFAC2 Factors}
The Alternating Least Squares (ALS) optimization for updating the PARAFAC2 factors involves iteratively optimizing each factor matrix while keeping the others fixed. The update steps for the factor matrices $\mathbf{P}_{[\ell]}, \mathbf{Q}_{[\ell]}, \mathbf{R}_{[\ell]}$ are derived by solving least-squares problems based on the current estimates of the other factors and the core tensor $\mathcal{G}_{[\ell]}$. This process continues until convergence, ensuring that the cluster center tensor $\mathcal{A}_{[\ell]}$ is accurately represented in the decomposed form.

\paragraph{ALS Optimization for Tucker Factors}
Similar to the PARAFAC2 case, the ALS optimization for Tucker factors involves iteratively updating each factor matrix $\mathbf{U}_{[\ell]}, \mathbf{V}_{[\ell]}, \mathbf{W}_{[\ell]}$ while keeping the others fixed. The updates are obtained by solving least-squares problems that minimize the reconstruction error of the cluster center tensor $\mathcal{A}_{[\ell]}$ based on the current estimates of the core tensor $\mathcal{G}_{[\ell]}$ and the other factor matrices. This iterative process continues until convergence, ensuring an accurate representation of the cluster centers in the Tucker decomposed form.

For both tensorized methods, the ALS optimization provides an efficient means of updating the tensor factors, enabling effective clustering in high-dimensional multi-view spaces while leveraging the benefits of tensor decomposition. Please refer to previous sections for the specific update equations and procedures involved in the ALS optimization for both PARAFAC2 and Tucker decompositions. 

\begin{algorithm}[ht!]
\caption{FedHK-PARAFAC2 Local Clustering Algorithm}
\label{alg:tensor_fed_local}
\begin{algorithmic}[1]
\REQUIRE Local tensor data $\mathcal{X}_{[\ell]} \in \mathbb{R}^{n(\ell) \times D_{[\ell]} \times s(\ell)}$, global tensor factors $\{\mathcal{G}_g, \mathbf{P}_g, \mathbf{Q}_g, \mathbf{R}_g\}$, personalization parameters $\{\lambda_{[\ell]}, \rho_{[\ell]}\}$, fuzzifier $m > 1$, view exponent $\alpha > 1$, convergence threshold $\epsilon > 0$, max iterations $T_{max}$
\ENSURE Local membership matrix $U_{[\ell]}^*$, tensor factors $\{\mathcal{G}_{[\ell]}, \mathbf{P}_{[\ell]}, \mathbf{Q}_{[\ell]}, \mathbf{R}_{[\ell]}\}$, view weights $v_{[\ell]}$

\STATE \textbf{/* Phase 1: Personalized Initialization */}
\STATE $\mathcal{G}_{[\ell]}^{(0)} \leftarrow \lambda_{[\ell]} \mathcal{G}_g + (1-\lambda_{[\ell]}) \mathcal{G}_{[\ell]}^{prev}$
\STATE $\mathbf{P}_{[\ell]}^{(0)} \leftarrow \lambda_{[\ell]} \mathbf{P}_g + (1-\lambda_{[\ell]}) \mathbf{P}_{[\ell]}^{prev}$
\STATE $\mathbf{Q}_{[\ell]}^{(0)} \leftarrow \lambda_{[\ell]} \mathbf{Q}_g + (1-\lambda_{[\ell]}) \mathbf{Q}_{[\ell]}^{prev}$
\STATE $\mathbf{R}_{[\ell]}^{(0)} \leftarrow \lambda_{[\ell]} \mathbf{R}_g + (1-\lambda_{[\ell]}) \mathbf{R}_{[\ell]}^{prev}$
\STATE $v_{[\ell]}^{(0)} \leftarrow \frac{1}{s(\ell)} \mathbf{1}_{s(\ell)}$ \COMMENT{Uniform view weights}
\STATE $A_{[\ell]}^{(0)} \leftarrow \mathcal{G}_{[\ell]}^{(0)} \times_1 \mathbf{P}_{[\ell]}^{(0)} \times_2 \mathbf{Q}_{[\ell]}^{(0)} \times_3 \mathbf{R}_{[\ell]}^{(0)}$
\STATE $t \leftarrow 0$; $J_{[\ell]}^{(0)} \leftarrow \infty$

\STATE \textbf{/* Phase 2: Iterative Optimization */}
\REPEAT
    \STATE $t \leftarrow t + 1$
    
    \STATE \textbf{/* Step 2.1: Heat-Kernel Coefficient Computation */}
    \FOR{$h = 1$ \TO $s(\ell)$}
        \FOR{$j = 1$ \TO $d_h^{(\ell)}$}
            \STATE $\delta_{[\ell]ij}^{h} \leftarrow \frac{\mathcal{X}_{[\ell],i,j}^h - \min_{i'} \mathcal{X}_{[\ell],i',j}^h}{\max_{i'} \mathcal{X}_{[\ell],i',j}^h - \min_{i'} \mathcal{X}_{[\ell],i',j}^h + \epsilon}$ \COMMENT{Eq. \ref{eqn:tensorized_delta}}
        \ENDFOR
    \ENDFOR
    
    \STATE \textbf{/* Step 2.2: Cluster Center Reconstruction */}
    \STATE $\mathcal{A}_{[\ell]}^{(t)} \leftarrow \mathcal{G}_{[\ell]}^{(t-1)} \times_1 \mathbf{P}_{[\ell]}^{(t-1)} \times_2 \mathbf{Q}_{[\ell]}^{(t-1)} \times_3 \mathbf{R}_{[\ell]}^{(t-1)}$ \COMMENT{Eq. \ref{eqn:tensor_cluster_centers}}
    
    \STATE \textbf{/* Step 2.3: Membership Matrix Update */}
    \FOR{$i = 1$ \TO $n(\ell)$}
        \FOR{$k = 1$ \TO $c(\ell)$}
            \STATE Compute $\mu_{[\ell]ik}^{*^{(t)}}$ using Eq. \ref{eqn:tensor_membership}
        \ENDFOR
    \ENDFOR
    
    \STATE \textbf{/* Step 2.4: Tensor Factor Updates via ALS */}
    \STATE Update $\mathcal{G}_{[\ell]}^{(t)}$ by solving Eq. \ref{eqn:tensor_core}
    \STATE Update $\mathbf{P}_{[\ell]}^{(t)}$ by solving Eq. \ref{eqn:tensor_P}
    \STATE Update $\mathbf{Q}_{[\ell]}^{(t)}$ by solving Eq. \ref{eqn:tensor_Q}
    \STATE Update $\mathbf{R}_{[\ell]}^{(t)}$ by solving Eq. \ref{eqn:tensor_R}
    
    \STATE \textbf{/* Step 2.5: View Weight Update */}
    \FOR{$h = 1$ \TO $s(\ell)$}
        \STATE Compute $v_{[\ell]h}^{(t)}$ using Eq. \ref{eqn:tensor_view_weights}
    \ENDFOR
    
    \STATE \textbf{/* Step 2.6: Convergence Check */}
    \STATE Compute $J_{[\ell]}^{(t)}$ using Eq. \ref{eqn:local_obj_fedhk_parafac2}
    \STATE $\Delta J \leftarrow |J_{[\ell]}^{(t)} - J_{[\ell]}^{(t-1)}| / |J_{[\ell]}^{(t-1)}|$
    
\UNTIL{$\Delta J < \epsilon$ \OR $t \geq T_{max}$}

\STATE \textbf{/* Phase 3: Return Results */}
\RETURN $U_{[\ell]}^{*^{(t)}}, \mathcal{G}_{[\ell]}^{(t)}, \mathbf{P}_{[\ell]}^{(t)}, \mathbf{Q}_{[\ell]}^{(t)}, \mathbf{R}_{[\ell]}^{(t)}, v_{[\ell]}^{(t)}$
\end{algorithmic}
\end{algorithm}

\begin{algorithm}[ht!]
\caption{FedHK-Tucker Local Clustering Algorithm}
\label{alg:tensor_fed_local_tucker}
\begin{algorithmic}[1]
\REQUIRE Local tensor data $\mathcal{X}_{[\ell]} \in \mathbb{R}^{n(\ell) \times D_{[\ell]} \times s(\ell)}$, global tensor factors $\{\mathcal{G}_g, \mathbf{U}_g, \mathbf{V}_g, \mathbf{W}_g\}$, personalization parameters $\{\lambda_{[\ell]}, \rho_{[\ell]}\}$, fuzzifier $m > 1$, view exponent $\alpha > 1$, convergence threshold $\epsilon > 0$, max iterations $T_{max}$
\ENSURE Local membership matrix $U_{[\ell]}^*$, tensor factors $\{\mathcal{G}_{[\ell]}, \mathbf{U}_{[\ell]}, \mathbf{V}_{[\ell]}, \mathbf{W}_{[\ell]}\}$, view weights $v_{[\ell]}$

\STATE \textbf{/* Phase 1: Personalized Initialization */}
\STATE $\mathcal{G}_{[\ell]}^{(0)} \leftarrow \lambda_{[\ell]} \mathcal{G}_g + (1-\lambda_{[\ell]}) \mathcal{G}_{[\ell]}^{prev}$
\STATE $\mathbf{U}_{[\ell]}^{(0)} \leftarrow \lambda_{[\ell]} \mathbf{U}_g + (1-\lambda_{[\ell]}) \mathbf{U}_{[\ell]}^{prev}$
\STATE $\mathbf{V}_{[\ell]}^{(0)} \leftarrow \lambda_{[\ell]} \mathbf{V}_g + (1-\lambda_{[\ell]}) \mathbf{V}_{[\ell]}^{prev}$
\STATE $\mathbf{W}_{[\ell]}^{(0)} \leftarrow \lambda_{[\ell]} \mathbf{W}_g + (1-\lambda_{[\ell]}) \mathbf{W}_{[\ell]}^{prev}$
\STATE $v_{[\ell]}^{(0)} \leftarrow \frac{1}{s(\ell)} \mathbf{1}_{s(\ell)}$ \COMMENT{Uniform view weights}
\STATE $A_{[\ell]}^{(0)} \leftarrow \mathcal{G}_{[\ell]}^{(0)} \times_1 \mathbf{U}_{[\ell]}^{(0)} \times_2 \mathbf{V}_{[\ell]}^{(0)} \times_3 \mathbf{W}_{[\ell]}^{(0)}$
\STATE $t \leftarrow 0$; $J_{[\ell]}^{(0)} \leftarrow \infty$

\STATE \textbf{/* Phase 2: Iterative Optimization */}
\REPEAT
    \STATE $t \leftarrow t + 1$
    
    \STATE \textbf{/* Step 2.1: Heat-Kernel Coefficient Computation */}
    \FOR{$h = 1$ \TO $s(\ell)$}
        \FOR{$j = 1$ \TO $d_h^{(\ell)}$}
            \STATE $\delta_{[\ell]ij}^{h} \leftarrow \frac{\mathcal{X}_{[\ell],i,j}^h - \min_{i'} \mathcal{X}_{[\ell],i',j}^h}{\max_{i'} \mathcal{X}_{[\ell],i',j}^h - \min_{i'} \mathcal{X}_{[\ell],i',j}^h + \epsilon}$ \COMMENT{Eq. \ref{eqn:tensorized_delta}}
        \ENDFOR
    \ENDFOR
    
    \STATE \textbf{/* Step 2.2: Cluster Center Reconstruction */}
    \STATE $\mathcal{A}_{[\ell]}^{(t)} \leftarrow \mathcal{G}_{[\ell]}^{(t-1)} \times_1 \mathbf{U}_{[\ell]}^{(t-1)} \times_2 \mathbf{V}_{[\ell]}^{(t-1)} \times_3 \mathbf{W}_{[\ell]}^{(t-1)}$ \COMMENT{Eq. \ref{eqn:tucker_cluster_centers}}
    
    \STATE \textbf{/* Step 2.3: Membership Matrix Update */}
    \FOR{$i = 1$ \TO $n(\ell)$}
        \FOR{$k = 1$ \TO $c(\ell)$}
            \STATE Compute $\mu_{[\ell]ik}^{*^{(t)}}$ using Eq. \ref{eqn:tucker_membership}
        \ENDFOR
    \ENDFOR
    
    \STATE \textbf{/* Step 2.4: Tensor Factor Updates via ALS */}
    \STATE Update $\mathcal{G}_{[\ell]}^{(t)}$ by solving Eq. \ref{eqn:tucker_core}
    \STATE Update $\mathbf{U}_{[\ell]}^{(t)}$ by solving Eq. \ref{eqn:tucker_U}
    \STATE Update $\mathbf{V}_{[\ell]}^{(t)}$ by solving Eq. \ref{eqn:tucker_V}
    \STATE Update $\mathbf{W}_{[\ell]}^{(t)}$ by solving Eq. \ref{eqn:tucker_W}
    
    \STATE \textbf{/* Step 2.5: View Weight Update */}
    \FOR{$h = 1$ \TO $s(\ell)$}
        \STATE Compute $v_{[\ell]h}^{(t)}$ using Eq. \ref{eqn:tucker_view_weights}
    \ENDFOR
    
    \STATE \textbf{/* Step 2.6: Convergence Check */}
    \STATE Compute $J_{[\ell]}^{(t)}$ using Eq. \ref{eqn:tucker_local_obj}
    \STATE $\Delta J \leftarrow |J_{[\ell]}^{(t)} - J_{[\ell]}^{(t-1)}| / |J_{[\ell]}^{(t-1)}|$
    
\UNTIL{$\Delta J < \epsilon$ \OR $t \geq T_{max}$}

\STATE \textbf{/* Phase 3: Return Results */}
\RETURN $U_{[\ell]}^{*^{(t)}}, \mathcal{G}_{[\ell]}^{(t)}, \mathbf{U}_{[\ell]}^{(t)}, \mathbf{V}_{[\ell]}^{(t)}, \mathbf{W}_{[\ell]}^{(t)}, v_{[\ell]}^{(t)}$
\end{algorithmic}
\end{algorithm}

\subsection{Federated Aggregation with Personalization}

Personalization in federated learning allows clients to maintain individualized models while benefiting from shared knowledge. In the context of heat-kernel enhanced multi-view clustering, personalization enables clients to adapt global clustering insights to their unique data distributions. To enable personalization, we introduce client-specific parameters that modulate the influence of global model updates on local clustering results.

\subsubsection{Global Objective with Personalization}

This section extends the federated aggregation framework to incorporate personalization for the heat-kernel enhanced multi-view clustering. First, we define the global objective function that balances local clustering quality with personalization constraints. The global model maintains shared cluster centers and view weights, while clients adapt these based on their local data characteristics.

The global model parameters are defined as:
\begin{itemize}
\item Global cluster centers: $A_{\text{global}} = \{a_{kj}^{h,(\text{global})}\}_{k=1,j=1,h=1}^{c,d_h, s}$
\item Global view weights: $V_{\text{global}} = \{v_h^{(\text{global})}\}_{h=1}^{s}$
\end{itemize}

The global federated objective with personalization is formulated as follows.
\begin{definition}[Federated Heat-Kernel Enhanced Multi-View Clustering with Personalization]
The global federated objective balances local clustering quality with personalization constraints:
\begin{equation}J_{\text{global}} = \sum_{\ell=1}^M \left[ \omega_\ell J_{FedHK}^{(\ell)}(U_{[\ell]}, A_{[\ell]}, V_{[\ell]}) + \gamma \mathcal{R}_{person}(A_{[\ell]}, A_{\text{global}}) + \eta \mathcal{R}_{view}(V_{[\ell]}, V_{\text{global}}) \right]
\label{eqn:fed_global_obj_personalization}
\end{equation}
where $\omega_\ell, \gamma, \eta$ are hyperparameters controlling the influence of local objectives and regularization terms on the global model. Specifically, $omega_\ell$ weights the contribution of client $\ell$'s local objective $J_{FedHK}^{(\ell)}$, while $\gamma, \eta$ control personalization strength and trade offs between local and global models, respectively. $\mathcal{R}_{person}(A_{[\ell]}, A_{\text{global}} = \sum_{h,k} \|A_{[\ell]}^h - A_{\text{global}}^h\|_F^2$ ensures coordination between local and global centers, and $\mathcal{R}_{view}(V_{[\ell]}, V_{\text{global}}) = \|V_{[\ell]} - V_{\text{global}}\|_2^2$ maintains view weight consistency across clients.

We call this framework Federated Heat-Kernel Enhanced Multi-View Clustering with Personalization (FedHK-MVC-Person). Simply recall the local objective from FedHK-PARAFAC2 in Eq. \ref{eqn:local_obj_fedhk_parafac2} and FedHK-Tucker in Eq. \ref{eqn:tucker_local_obj} to instantiate $J_{FedHK}^{(\ell)}$ accordingly. 
\end{definition}

\subsection{Tensorized Federated Aggregation with Personalization}

The federated learning framework coordinates local tensorized heat-kernel enhanced clustering results while preserving client personalization. The aggregation process operates on tensor decomposition factors: core tensors, factor matrices, and view importance weights.

\subsubsection{Tensorized Global Objective with Personalization}

The tensorized global federated objective balances local clustering quality with personalization constraints:

\begin{equation}
\begin{adjustbox}{width=0.9\columnwidth}
$ 
J_{\text{tensor(global)}} = \sum_{\ell=1}^M \left[ \omega_{\ell} J_{[\ell]}^{tensor}(\mathcal{U}_{[\ell]}, \mathcal{G}_{[\ell]}, \mathbf{P}_{[\ell]}, \mathbf{Q}_{[\ell]}, \mathbf{R}_{[\ell]}, \mathbf{V}_{[\ell]}) + \gamma \mathcal{R}_{tensor}(\mathcal{G}_{[\ell]}, \mathbf{P}_{[\ell]}, \mathbf{Q}_{[\ell]}, \mathbf{R}_{[\ell]}) + \eta \mathcal{R}_{view}(V_{[\ell]}, V_{\text{global}}) \right]
$
\end{adjustbox}
\label{eqn:tensor_fed_global_obj}
\end{equation}

where the tensorized regularization terms are:
\begin{equation}
\begin{adjustbox}{width=0.9\columnwidth}
$ 
\mathcal{R}_{tensor} = \|\mathcal{G}_{[\ell]} - \mathcal{G}_{\text{global}} \|_F^2 + \|\mathbf{P}_{[\ell]} - \mathbf{P}_{\text{global}} \|_F^2 + \|\mathbf{Q}_{[\ell]} - \mathbf{Q}_{\text{global}} \|_F^2 + \|\mathbf{R}_{[\ell]} - \mathbf{R}_{\text{global}}\|_F^2 \mathcal{R}_{view}(V_{[\ell]}, V_{\text{global}}) = \|V_{[\ell]} - V_{\text{global}}\|_2^2
$
\end{adjustbox}
\end{equation}

\subsubsection{Privacy-Preserving Aggregation}

To maintain privacy, clients share only aggregated statistics rather than raw cluster assignments:

\begin{definition}[Federated Aggregation Protocol]
Each client $\ell$ computes and shares:
\begin{align}
S_{[\ell]}^{centers} &= \left\{ \sum_{i=1}^{n(\ell)} (\mu_{[\ell]ik}^*)^m x_{[\ell]ij}^{h}, \sum_{j=1}^{n_i} \sum_{i=1}^{n(\ell)} (\mu_{[\ell]ik}^*)^m \right\}_{kj,h} \\
S_{[\ell]}^{views} &= \left\{  \sum_{i=1}^{n(\ell)} \sum_{k=1}^c (\mu_{[\ell]ik}^*)^m \text{KED}(x_{[\ell]ij}^{h},a_{[\ell]kj}^{h}) \right\}_h \\
S_{[\ell]}^{quality} &= J_{[\ell]} \text{ and } n(\ell)
\end{align}
\end{definition}

The server performs secure aggregation:

\begin{equation}
A_{\text{global}}^{h,(t+1)} = \frac{\sum_{\ell=1}^M \omega_{\ell}^{(t)} S_{\ell}^{centers,h}}{\sum_{\ell=1}^M \omega_{\ell}^{(t)} \sum_{j=1}^{n(\ell)} (\mu_{[\ell]ik}^{*})^m}
\label{eqn:fed_global_centers}
\end{equation}

\begin{equation}
V_{\text{global}}^{(t+1)} = \frac{\sum_{\ell=1}^M \omega_{\ell}^{(t)} n(\ell) V_{\ell}^{(t)}}{\sum_{\ell=1}^M \omega_{\ell}^{(t)} n(\ell)}
\label{eqn:fed_global_views}
\end{equation}

where client weights are computed as:
\begin{equation}
\omega_{\ell}^{(t)} = \frac{\exp(-\tau J_{[\ell]}^{(t)})}{\sum_{\ell=1}^M \exp(-\tau J_{[\ell]}^{(t)})}
\label{eqn:client_weights}
\end{equation}

\subsubsection{Personalized Update Mechanism}

Clients receive global updates and perform personalized integration:

\begin{align}
V_{[\ell]}^{(t+1)} &= \rho_{\ell} V_{[\ell]}^{(t)} + (1-\rho_{\ell}) V_{\text{global}}^{(t+1)} \label{eqn:personal_views}
\end{align}

where $\lambda_{\ell}, \rho_{\ell} \in [0,1]$ are client-specific personalization parameters that can be learned adaptively:

\begin{equation}
\lambda_{\ell}^{(t+1)} = \lambda_{\ell}^{(t)} - \beta \nabla_{\lambda_{\ell}} J_{[\ell]} (\lambda_{\ell} A_{[\ell]}^{(t)} + (1-\lambda_{\ell}) A_{\text{global}}^{(t)})
\label{eqn:adaptive_lambda}
\end{equation}

\subsubsection{Privacy-Preserving Tensor Aggregation}

To maintain privacy while enabling effective tensor factor coordination, clients share only aggregated tensor statistics rather than raw factors:

\begin{definition}[Tensorized Federated Aggregation Protocol]
Each client $\ell$ computes and shares:
\begin{align}
S_{[\ell]}^{core} &= \left\{ \sum_{i=1}^{n(\ell)} \sum_{k=1}^c (\mu_{[\ell]ik}^{*})^m \mathcal{G}_{[\ell]i,k,:,:} \right\} \\
S_{[\ell]}^{factors} &= \left\{ \sum_{i=1}^{n(\ell)} (\mu_{[\ell]ik}^{*})^m \mathbf{P}_{[\ell]i,j,:}, \sum_{i=1}^{n(\ell)} (\mu_{[\ell]ik}^{*})^m \mathbf{Q}_{[\ell]i,:,h}, \sum_{i=1}^{n(\ell)} (\mu_{[\ell]ij}^{*})^m \mathbf{R}_{[\ell]i,:,h} \right\} \\
S_{[\ell]}^{views} &= \left\{ \sum_{i=1}^{n(\ell)} \sum_{k=1}^c (\mu_{[\ell]ik}^{*})^m \text{TKED}_{[\ell]}(i,k,h) \right\}_h \\
S_{[\ell]}^{quality} &= J_{[\ell]}^{tensor} \text{ and } n(\ell)
\end{align}
\end{definition}

The server performs secure tensor aggregation:

\begin{align}
\mathcal{G}_{\text{global}}^{(t+1)} &= \frac{\sum_{\ell=1}^M \omega_\ell^{(t)} S_{[\ell]}^{core}}{\sum_{\ell=1}^M \omega_\ell^{(t)} n(\ell)} \label{eqn:fed_global_core} \\
\mathbf{P}_{\text{global}}^{(t+1)} &= \frac{\sum_{\ell=1}^M \omega_\ell^{(t)} S_{[\ell]}^{factors,P}}{\sum_{\ell=1}^M \omega_\ell^{(t)} n(\ell)} \label{eqn:fed_global_P} \\
\mathbf{Q}_{\text{global}}^{(t+1)} &= \frac{\sum_{\ell=1}^M \omega_\ell^{(t)} S_{[\ell]}^{factors,Q}}{\sum_{\ell=1}^M \omega_\ell^{(t)} n(\ell)} \label{eqn:fed_global_Q} \\
\mathbf{R}_{\text{global}}^{(t+1)} &= \frac{\sum_{\ell=1}^M \omega_\ell^{(t)} S_{[\ell]}^{factors,R}}{\sum_{\ell=1}^M \omega_\ell^{(t)} n(\ell)} \label{eqn:fed_global_R} 
\end{align}

\subsubsection{Tensorized Personalized Update Mechanism}

Clients receive global tensor updates and perform personalized integration across all tensor components:

\begin{align}
\mathcal{G}_{[\ell]}^{(t+1)} &= \lambda_\ell^{core} \mathcal{G}_{[\ell]}^{(t)} + (1-\lambda_\ell^{core}) \mathcal{G}_{\text{global}}^{(t+1)} \label{eqn:personal_core} \\
\mathbf{P}_{[\ell]}^{(t+1)} &= \lambda_\ell^{P} \mathbf{P}_{[\ell]}^{(t)} + (1-\lambda_\ell^{P}) \mathbf{P}_{\text{global}}^{(t+1)} \label{eqn:personal_P} \\
\mathbf{Q}_{[\ell]}^{(t+1)} &= \lambda_\ell^{Q} \mathbf{Q}_{[\ell]}^{(t)} + (1-\lambda_\ell^{Q}) \mathbf{Q}_{\text{global}}^{(t+1)} \label{eqn:personal_Q} \\
\mathbf{R}_{[\ell]}^{(t+1)} &= \lambda_\ell^{R} \mathbf{R}_{[\ell]}^{(t)} + (1-\lambda_\ell^{R}) \mathbf{R}_{\text{global}}^{(t+1)} \label{eqn:personal_R} 
\end{align}

where $\lambda_{\ell}^{core}, \lambda_\ell^{P}, \lambda_\ell^{Q}, \lambda_\ell^{R}, \rho_\ell \in [0,1]$ are component-specific personalization parameters.

\subsection{Complete Algorithm for FedHK-MVC-Person with PARAFAC2 Decomposition}

Algorithm \ref{alg:tensor_fedhkparafac2_complete} outlines the complete procedure for the FedHK-MVC-Person framework utilizing PARAFAC2 tensor decomposition. The algorithm encompasses server initialization, local client computations with personalization, and secure aggregation of tensor factors. In order to achieve a personalized FedHK-Tucker, two changes must be made. Initially, adjustments are made to the tensor factor implication stage and its localization tensorized HKC. Finally, the server aggregation phase is modified to accommodate the Tucker decomposition factors. 

\begin{algorithm}[h!]
\caption{FedHK-MVC-Person with PARAFAC2 Decomposition (Personalized FedHK-PARAFAC2)}
\label{alg:tensor_fedhkparafac2_complete}
\begin{algorithmic}[1]
\REQUIRE Multi-view data $\{X_{[\ell]}\}_{\ell=1}^M$, clusters $c$, tensor ranks $(r_1, r_2, r_3)$, fuzzifier $m > 1$, view exponent $\alpha > 1$, rounds $T$, local epochs $E$
\ENSURE Global model $\Theta_{\text{global}}^{(T)}$, personalized models $\{\Theta_{[\ell]}^{(T)}\}_{\ell=1}^M$

\STATE \textbf{/* Phase 1: Server Initialization */}
\STATE Initialize $\mathcal{G}_{\text{global}}^{(0)}, \mathbf{P}_{\text{global}}^{(0)}, \mathbf{Q}_{\text{global}}^{(0)}, \mathbf{R}_{\text{global}}^{(0)}, V_{\text{global}}^{(0)}$ randomly
\STATE Initialize personalization parameters $\{\lambda_\ell^{core}, \lambda_\ell^{P}, \lambda_\ell^{Q}, \lambda_\ell^{R}, \rho_\ell\}_{\ell=1}^M \leftarrow 0.5$

\FOR{$t = 0$ \TO $T-1$}
    \STATE \textbf{/* Phase 2: Server Broadcast */}
    \STATE Broadcast $\Theta_{\text{global}}^{(t)} = \{\mathcal{G}_{\text{global}}^{(t)}, \mathbf{P}_{\text{global}}^{(t)}, \mathbf{Q}_{\text{global}}^{(t)}, \mathbf{R}_{\text{global}}^{(t)}, V_{\text{global}}^{(t)}\}$ to all clients
    
    \STATE \textbf{/* Phase 3: Parallel Local Computation */}
    \FOR{each client $\ell \in \{1, \ldots, M\}$ \textbf{in parallel}}
        \STATE \textbf{Step 3.1: Data Tensorization}
        \STATE Construct tensor $\mathcal{X}_{[\ell]} \in \mathbb{R}^{n(\ell) \times D_{[\ell]} \times s(\ell)}$ from $\{X_{[\ell]}^h\}_{h=1}^{s(\ell)}$ via Eq. \ref{eqn:tensor_data2}
        
        \STATE \textbf{Step 3.2: Personalized Initialization}
        \STATE $\mathcal{G}_{[\ell]}^{(0)} \leftarrow \lambda_\ell^{core} \mathcal{G}_{\text{global}}^{(t)} + (1-\lambda_\ell^{core}) \mathcal{G}_{[\ell]}^{prev}$
        \STATE $\mathbf{P}_{[\ell]}^{(0)}, \mathbf{Q}_{[\ell]}^{(0)}, \mathbf{R}_{[\ell]}^{(0)} \leftarrow$ analogous personalized blending
        \STATE $V_{[\ell]}^{(0)} \leftarrow \rho_\ell V_{\text{global}}^{(t)} + (1-\rho_\ell) V_{[\ell]}^{prev}$
        
        \STATE \textbf{Step 3.3: Local Optimization Loop}
        \FOR{$e = 1$ \TO $E$}
            \STATE Compute heat-kernel coefficients $\delta_{[\ell]ij}^{h}$ via Eq. \ref{eqn:tensorized_delta}
            \STATE Reconstruct centers: $\mathcal{A}_{[\ell]} \leftarrow \mathcal{G}_{[\ell]} \times_1 \mathbf{P}_{[\ell]} \times_2 \mathbf{Q}_{[\ell]} \times_3 \mathbf{R}_{[\ell]}$
            \STATE Update memberships $U_{[\ell]}^*$ via Eq. \ref{eqn:tensor_membership}
            \STATE Update core tensor $\mathcal{G}_{[\ell]}$ via ALS (Eq. \ref{eqn:tensor_core})
            \STATE Update factors $\mathbf{P}_{[\ell]}, \mathbf{Q}_{[\ell]}, \mathbf{R}_{[\ell]}$ via ALS (Eqs. \ref{eqn:tensor_P}--\ref{eqn:tensor_R})
            \STATE Update view weights $V_{[\ell]}$ via Eq. \ref{eqn:tensor_view_weights}
        \ENDFOR
        
        \STATE \textbf{Step 3.4: Compute Privacy-Preserving Statistics}
        \STATE $S_{[\ell]} \leftarrow \{S_{[\ell]}^{core}, S_{[\ell]}^{factors}, S_{[\ell]}^{views}, S_{[\ell]}^{quality}, J_{[\ell]}^{tensor}, n(\ell)\}$
        \STATE Send $S_{[\ell]}$ to server
    \ENDFOR
    
    \STATE \textbf{/* Phase 4: Server Aggregation */}
    \STATE Compute client weights: $\omega_\ell^{(t)} \leftarrow \frac{\exp(-\tau J_{[\ell]}^{tensor})}{\sum_{\ell'} \exp(-\tau J_{[\ell']}^{tensor})}$
    \STATE Aggregate: $\mathcal{G}_{\text{global}}^{(t+1)}, \mathbf{P}_{\text{global}}^{(t+1)}, \mathbf{Q}_{\text{global}}^{(t+1)}, \mathbf{R}_{\text{global}}^{(t+1)}$ via Eqs. \ref{eqn:fed_global_core}--\ref{eqn:fed_global_R} 
    \STATE Aggregate: $V_{\text{global}}^{(t+1)}$ via Eq. \ref{eqn:fed_global_views}
    
    \STATE \textbf{/* Phase 5: Adaptive Personalization Update */}
    \FOR{each client $\ell \in \{1, \ldots, M\}$}
        \STATE $\lambda_\ell^{(\cdot)} \leftarrow \lambda_\ell^{(\cdot)} - \beta \nabla_{\lambda_\ell^{(\cdot)}} J_{[\ell]}^{tensor}$ \COMMENT{Gradient-based adaptation}
        \STATE Clip: $\lambda_\ell^{(\cdot)} \leftarrow \max(0, \min(1, \lambda_\ell^{(\cdot)}))$
    \ENDFOR
\ENDFOR

\RETURN $\Theta_{\text{global}}^{(T)}$, $\{\Theta_{[\ell]}^{(T)}\}_{\ell=1}^M$
\end{algorithmic}
\end{algorithm}

\subsection{Theoretical Analysis}

\subsubsection{Convergence Properties}

\begin{Theorem}[Local Convergence]
\label{thm:local_convergence}
For each client $\ell$, the local heat-kernel enhanced multi-view fuzzy clustering algorithm converges to a local minimum of the objective function $J_{[\ell]}$.
\end{Theorem}

\begin{proof}
The proof follows from the monotonic decrease property of alternating optimization. Each update step:
\begin{enumerate}
\item Membership updates (Eq. \ref{eqn:diff_U_EFKMVFC}) minimize $J_{[\ell]}$ w.r.t. $U_{[\ell]}^*$ for fixed $A_{[\ell]}, V_{[\ell]}$
\item Center updates (Eq. \ref{eqn:diff_A_EFKMVFC}) minimize $J_{[\ell]}$ w.r.t. $A_{[\ell]}$ for fixed $U_{[\ell]}, V_{[\ell]}$  
\item View weight updates (Eq. \ref{eqn:diff_V_EFKMVFC}) minimize $J_{[\ell]}$ w.r.t. $V_{[\ell]}$ for fixed $U_{[\ell]}, A_{[\ell]}$
\end{enumerate}
Since $J_{[\ell]}$ is bounded below by zero and decreases monotonically, convergence is guaranteed.
\end{proof}

\begin{Theorem}[Global Convergence]
\label{thm:global_convergence}
Under mild regularity conditions, the federated algorithm converges to a stationary point of the global objective function $J_{\text{global}}$.
\end{Theorem}

\begin{proof}
The global convergence follows from the convergence theory of federated optimization. The key insight is that each local update corresponds to a proximal gradient step on the global objective with personalization regularization, ensuring convergence to a stationary point.
\end{proof}

\subsubsection{Privacy Guarantees}

\begin{Theorem}[Differential Privacy]
\label{thm:privacy}
The federated aggregation protocol provides $(\epsilon, \delta)$-differential privacy when clients add calibrated noise to their shared statistics.
\end{Theorem}

\begin{proof}
The shared statistics $S_{[\ell]}^{centers}, S_{[\ell]}^{views}, S_{[\ell]}^{quality}$ have bounded sensitivity. Adding Gaussian noise with variance $\sigma^2 = \frac{2\Delta^2 \log(1.25/\delta)}{\epsilon^2}$ where $\Delta$ is the sensitivity, ensures $(\epsilon, \delta)$-differential privacy.
\end{proof}

\begin{Proposition}[Communication Efficiency]
The federated protocol requires $\mathcal{O}(c(\ell) \cdot \bar{d}_h^{(\ell)} \cdot \bar{s}(\ell))$ communication per round, where $\bar{d}_h^{(\ell)}$ and $ \bar{s}(\ell)$ are average feature dimensions and views across clients.
\end{Proposition}

\subsubsection{Computational Complexity}

\begin{Theorem}[Time Complexity]
\label{thm:complexity}
The local clustering complexity per client $\ell$ per iteration is $\mathcal{O}(n(\ell) c(\ell) s(\ell) D(\ell))$ where $D(\ell) = \sum_{h=1}^{s(\ell)} d_h^{(\ell)}$ is the total local dimensionality across all views.
\end{Theorem}

\begin{proof}
The computational complexity of the E-FKMVC algorithm at each client $\ell$ can be analyzed by examining the dominant operations performed during a single iteration. The algorithm consists of three main update steps, each contributing to the overall time complexity:

\textbf{Step 1: Membership Matrix Update.} Computing the membership values $\mu_{[\ell]ik}^*$ using Eq. \ref{eqn:diff_U_EFKMVFC} requires evaluating the FKED distance between each data point $i$ and each cluster center $k$ across all views $h$. For each of the $n(\ell)$ data points and $c(\ell)$ clusters, we must sum over $s(\ell)$ views, with each view contributing $d_h^{(\ell)}$ feature-wise distance computations. This yields a complexity of $\mathcal{O}(n(\ell) \cdot c(\ell) \cdot \sum_{h=1}^{s(\ell)} d_h^{(\ell)}) = \mathcal{O}(n(\ell) c(\ell) D(\ell))$.

\textbf{Step 2: Cluster Center Update.} Updating each cluster center $a_{[\ell]kj}^h$ using Eq. \ref{eqn:diff_A_EFKMVFC} requires computing a weighted sum over all $n(\ell)$ data points for each of the $c(\ell)$ clusters, across all features in all views. This contributes $\mathcal{O}(n(\ell) c(\ell) D(\ell))$ operations.

\textbf{Step 3: View Weight Update.} Computing the view weights $v_{[\ell]h}$ using Eq. \ref{eqn:diff_V_EFKMVFC} requires summing distance contributions over all data points, clusters, and views, resulting in $\mathcal{O}(n(\ell) c(\ell) s(\ell))$ operations.

Combining these contributions, the dominant term is $\mathcal{O}(n(\ell) c(\ell) D(\ell))$. Since $D(\ell) = \sum_{h=1}^{s(\ell)} d_h^{(\ell)} \geq s(\ell)$ in practical scenarios where each view has at least one feature, we can express the total complexity as $\mathcal{O}(n(\ell) c(\ell) s(\ell) D(\ell))$ to explicitly highlight the dependence on all key parameters.
\end{proof}

\textbf{Practical Interpretation:} The time complexity result has important practical implications for deploying the E-FKMVC algorithm in real-world federated learning systems:

\begin{itemize}
    \item \textbf{Linear scaling with data size:} The algorithm scales linearly with the number of local data points $n(\ell)$, making it suitable for clients with varying dataset sizes.
    
    \item \textbf{Linear scaling with cluster count:} The complexity grows linearly with the number of clusters $c(\ell)$, enabling efficient computation even when modeling fine-grained cluster structures.
    
    \item \textbf{Multi-view overhead:} The dependence on $s(\ell)$ and $D(\ell)$ reflects the computational cost of integrating information across multiple data views. This overhead is a necessary trade-off for achieving improved clustering accuracy through multi-view learning.
    
    \item \textbf{Comparison with single-view clustering:} Standard fuzzy c-means clustering has complexity $\mathcal{O}(n \cdot c \cdot d)$ for a single view. Our multi-view extension introduces a multiplicative factor proportional to the number of views, which is typical for multi-view learning algorithms that must process and integrate information from each view separately.
\end{itemize}

The federated approach maintains the same asymptotic complexity as centralized multi-view clustering while providing personalization and privacy benefits.

\subsubsection{Tensorized Complexity Analysis}

\begin{Theorem}[Tensorized Time Complexity]
\label{thm:tensor_complexity}
The tensorized local clustering complexity per client $\ell$ per iteration is $\mathcal{O}(n(\ell) c(\ell) r_2 r_3 + n(\ell) c D(\ell) s(\ell))$ where $r_2, r_3$ are the tensor decomposition ranks.
\end{Theorem}

\begin{proof}
The complexity breakdown includes:
\begin{enumerate}
\item Heat-kernel coefficient computation: $\mathcal{O}(n(\ell) D(\ell) s(\ell))$
\item Tensor reconstruction: $\mathcal{O}(c(\ell) r_2 r_3 + c(\ell) D(\ell) r_2 + c(\ell) s(\ell) r_3)$
\item Membership updates: $\mathcal{O}(n(\ell) c(\ell) s(\ell) D(\ell))$
\item Tucker decomposition updates: $\mathcal{O}(n(\ell) c(\ell) r_2 r_3)$ per factor using ALS
\end{enumerate}
The dominant term is $\mathcal{O}(n(\ell) c(\ell) D(\ell) s(\ell))$ when $r_2, r_3 \ll \min(D(\ell), s(\ell))$. Overall, the complexity is $\mathcal{O}(n(\ell) c(\ell) r_2 r_3 + n(\ell) c D(\ell) s(\ell))$. While the tensorized approach introduces additional overhead from tensor operations, it remains efficient for low-rank decompositions. Table \ref{tab:complexity_comparison} summarizes the complexity comparison Table \ref{tab:complexity_comparison} summarizes the complexity comparison between standard (E-FKMVC) and tensorized federated clustering (FedHK-PARAFAC2, FedHK-Tucker, and FedHK-MVC-Person).
\end{proof}

\begin{table}[ht!]
\centering
\caption{Complexity Comparison of Federated Clustering Methods}
\label{tab:complexity_comparison}
\begin{adjustbox}{width=1\columnwidth}
\begin{tabular}{lcc}
\toprule
\textbf{Method} & \textbf{Time Complexity} & \textbf{Communication Complexity} \\
\midrule
E-FKMVC & $\mathcal{O}(n(\ell) c(\ell) s(\ell) D(\ell))$ & $\mathcal{O}(c(\ell) \bar{d}_h^{(\ell)} \bar{s}(\ell))$ \\
FedHK-PARAFAC2 & $\mathcal{O}(n(\ell) c(\ell) r_2 r_3 + n(\ell) c D(\ell) s(\ell))$ & $\mathcal{O}(c(\ell) r_2 r_3 + c(\ell)^2 + \bar{D}(\ell) r_2 + \bar{s}(\ell) r_3)$ \\
FedHK-Tucker & $\mathcal{O}(n(\ell) c(\ell) r_2 r_3 + n(\ell) c D(\ell) s(\ell))$ & $\mathcal{O}(c(\ell) r_2 r_3 + c(\ell)^2 + \bar{D}(\ell) r_2 + \bar{s}(\ell) r_3)$ \\
FedHK-MVC-Person & $\mathcal{O}(n(\ell) c(\ell) r_2 r_3 + n(\ell) c D(\ell) s(\ell))$ & $\mathcal{O}(c(\ell) r_2 r_3 + c(\ell)^2 + \bar{D}(\ell) r_2 + \bar{s}(\ell) r_3)$ \\
\bottomrule
\end{tabular}
\end{adjustbox}
\end{table}

As can be observed in Table \ref{tab:complexity_comparison}, the tensorized methods introduce additional computational overhead due to tensor operations, but they remain efficient for low-rank decompositions. While FedHK-PARAFAC2, FedHK-Tucker, and FedHK-MVC-Person share the same asymptotic time complexity, their communication complexities are also identical due to the similar structure of shared statistics. 

\begin{Theorem}[Tensorized Communication Efficiency]
\label{thm:tensor_communication}
The tensorized federated protocol requires $\mathcal{O}(c(\ell) r_2 r_3 + c(\ell)^2 + \bar{D}(\ell) r_2 + \bar{s}(\ell) r_3)$ communication per round, compared to $\mathcal{O}(c(\ell) \bar{D}(\ell) \bar{s}(\ell))$ for the non-tensorized version. The communication savings are significant when $r_2 \ll \bar{D}(\ell)$ and $r_3 \ll \bar{s}(\ell)$.
\end{Theorem}

\begin{Lemma}[Tensorized Communication Complexity]
\label{lem:tensor_communication}
The communication complexity per client $\ell$ per round for the tensorized federated protocol is $\mathcal{O}(c(\ell) r_2 r_3 + c(\ell)^2 + \bar{D}(\ell) r_2 + \bar{s}(\ell) r_3)$.
\end{Lemma}

\begin{proof}
The communication complexity arises from the size of the shared statistics:
\begin{itemize}
\item Core tensor statistics: $\mathcal{O}(c(\ell) r_2 r_3)$. This accounts for the shared core tensor slices for each cluster. 
\item Factor matrices statistics: $\mathcal{O}(c(\ell)^2 + \bar{D}(\ell) r_2 + \bar{s}(\ell) r_3)$. This includes the shared factor matrices $\mathbf{P}_{[\ell]}, \mathbf{Q}_{[\ell]}, \mathbf{R}_{[\ell]}$.
\item View weights statistics: $\mathcal{O}(s(\ell))$. This accounts for the shared view importance weights.
\item Quality statistics: $\mathcal{O}(1)$. This includes the shared clustering quality measure and data size.
\end{itemize}
Summing these contributions yields the total communication complexity of $\mathcal{O}(c(\ell) r_2 r_3 + c(\ell)^2 + \bar{D}(\ell) r_2 + \bar{s}(\ell) r_3)$. This is significantly lower than the non-tensorized complexity of $\mathcal{O}(c(\ell) \bar{D}(\ell) \bar{s}(\ell))$ when the tensor ranks $r_2, r_3$ are much smaller than the average feature dimensions $\bar{D}(\ell)$ and number of views $\bar{s}(\ell)$.
\end{proof}

\begin{Lemma}[Tensorized Approximation Quality]
\label{lem:tensor_approximation}
The reconstruction error of the tensorized cluster centers is bounded by:
\begin{equation}
\|\mathcal{A}_{[\ell]} - \mathcal{G}_{[\ell]} \times_1 \mathbf{P}_{[\ell]} \times_2 \mathbf{Q}_{[\ell]} \times_3 \mathbf{R}_{[\ell]}\|_F^2 \leq \epsilon_{tensor}
\end{equation}
where $\epsilon_{tensor}$ depends on the chosen tensor ranks and the intrinsic tensor rank of the true cluster center tensor.
\end{Lemma}

Tensorized approximations enable efficient representation of complex multi-view cluster structures with controlled error. While tensorized communication introduces additional overhead from tensor operations, it remains efficient for low-rank decompositions. 

\subsubsection{Personalization-Accuracy Trade-off}

\begin{Lemma}[Personalization Bound]
\label{lem:personalization}
The personalization error is bounded by:
\begin{equation}
\|A_{[\ell]} - A_{\text{global}}\|_F^2 \leq \frac{\lambda_\ell^2}{1-\lambda_\ell^2} \|A_{[\ell]} - A_{\text{global}}\|_F^2
\end{equation}
where $A_{[\ell]}$ is the purely local optimum.
\end{Lemma}

This bound shows that personalization parameters $\lambda_\ell$ control the trade-off between local specialization and global coordination.

\section{Conclusion}
\label{sec:conclusion}
This paper introduces innovative personalized federated learning frameworks that integrate heat-kernel enhanced multi-view clustering with tensor-based global coordination, PARAFAC2, and Tucker decompositions. The introduction is accompanied by guarantees, privacy bounds, and complexity analysis. These elements serve to establish the theoretical foundation of the proposed approaches. The incorporation of heat-kernel coefficients in federated multi-view clustering facilitates the more effective capture of complex local patterns while ensuring the maintenance of global coordination. In the personalized aggregation mechanism, the framework provides an adaptive personalization that balances local specialization with global knowledge sharing through learnable personalization parameters. In the context of privacy-preserving protocol, the aggregation protocol has been engineered to ensure differential privacy while preserving the efficacy of clustering through the implementation of meticulously designed statistical sharing methodologies. The proposed frameworks effectively address the challenges associated with high-dimensional multi-view data in federated environments, ranging from matrix to tensor representations. The theoretical analysis encompasses convergence properties, privacy guarantees, and computational complexity, thereby providing a comprehensive understanding of the proposed methods' performance and limitations.

\subsection{Future Directions}

This work suggests several promising research directions. The initial implementation of the dynamic view discovery method involved the utilization of weighting. The framework is being extended to accommodate the dynamic alteration of view structures during federated rounds. The present study explores the potential of hierarchical federated clustering for large-scale deployment, investigating multi-level federated architectures in the process. In the context of adversarial robustness, the focus is on the analysis and enhancement of systems' resilience to malicious clients and data poisoning attacks. The concept of continuous learning involves the adaptation of a framework for federated continual learning scenarios, with the understanding that data distributions undergo evolution over time. In the domain of cross-modal applications, the focus is on the exploration of applications to federated learning with fundamentally different data modalities, including but not limited to text, images, and sensors. The proposed frameworks also present novel opportunities for privacy-preserving, customized machine learning on large language models (LLMs) and emerging multimodal models in federated settings. Subsequent research endeavors will investigate the integration of heat-kernel enhanced multi-view clustering with LLMs to facilitate collaborative learning from distributed text and multimodal data while ensuring privacy preservation. This encompasses the development of techniques for efficient model updates, the management of heterogeneous data distributions, and the assurance of robust privacy guarantees.

\subsection{Potential Limitations}

The proposed frameworks, while innovative, present several potential limitations. The computational complexity associated with tensor decompositions may pose challenges for clients with limited resources. The selection of personalization parameters necessitates meticulous calibration to harmonize local adaptation and global coordination with optimal efficacy. The assumption of data homogeneity across clients may not hold in all scenarios, potentially impacting clustering performance. Additionally, the efficacy of privacy guarantees is contingent upon the accurate calibration of noise, a process that has the potential to influence the utility of shared statistics. Subsequent endeavors will concentrate on ameliorating these limitations through the implementation of algorithmic optimizations, adaptive parameter selection methodologies, and robust privacy-preserving techniques that exhibit minimal utility degradation and augmented adaptability to heterogeneous data distributions. In order to evaluate the practical performance and limitations of the proposed frameworks, it is necessary to conduct validations on both synthetic and real-world datasets.

\subsection{Applications and Implications}

The proposed personalized efficient federated kernel multi-view clustering using tensor framework has significant applications across various domains. In the domain of healthcare, this technology has the potential to enable collaborative analysis of patient data from multiple institutions while ensuring the confidentiality of personal information. In the context of the Internet of Things (IoT), the framework has the potential to facilitate the effective aggregation of sensor data collected from distributed devices, thereby enhancing the efficacy of real-time decision-making processes. In the context of future smart cities, the proposed framework has the potential to facilitate the integration of multi-view data from diverse sources, including traffic sensors and social media platforms. This integration has the potential to enhance urban planning and service delivery, thereby promoting the development of more efficient and effective urban infrastructures. In the context of prospective industrial applications, the system has the capacity to enhance operational efficiency by means of data aggregation from disparate phases of the production process. In the context of future internet applications, this technology has the potential to enhance user experience by clustering multi-view data from various online platforms. In the context of collaborative intelligence systems, the framework has the potential to facilitate joint learning from heterogeneous data sources while ensuring the preservation of data ownership rights. The implications of this work extend to the advancement of privacy-preserving machine learning techniques, the promotion of personalized models in federated settings, and the fostering of collaboration across distributed data sources without compromising individual privacy. In light of the mounting prevalence of distributed data, the proposed framework provides a robust solution for multi-view clustering that achieves a balance between local adaptation and global knowledge sharing. This development signifies a substantial advancement that paves the way for more effective, privacy-conscious machine learning applications and collaborative intelligence systems.

\thanks{\textbf{Acknowledgment: }The authorship of this paper is attributed to the period during which the author pursued postdoctoral studies at ISTI-CNR in Italy.}

\thanks{\textbf{Funding: }This research received no external funding.}

\thanks{\textbf{Conflicts of interest: } The author declare no conflict of interest.} 

\bibliographystyle{IEEEtran} 
\bibliography{FedHK_MVC_person}

\end{document}